\documentclass[sigconf]{acmart}

\usepackage{booktabs}

\usepackage{comment}
\usepackage{multirow}
\usepackage{paralist}
\usepackage{graphicx}
\usepackage{subcaption}
\usepackage{rotating}
\usepackage{array}
\usepackage{balance}

\usepackage{changes}
\colorlet{Changes@Color}{red}

\graphicspath{{figures/}}

\setcopyright{acmlicensed}

\acmDOI{10.1145/3204493.3204545}

\acmISBN{978-1-4503-5706-7/18/06}

\acmConference[ETRA '18]{ETRA '18: 2018 Symposium on Eye Tracking Research and Applications}{June 14--17, 2018}{Warsaw, Poland}
\acmBooktitle{ETRA '18: ETRA '18: 2018 Symposium on Eye Tracking Research and Applications, June 14--17, 2018, Warsaw, Poland}
\acmYear{2018}
\copyrightyear{2018}

\acmPrice{15.00}

\acmSubmissionID{1116}

\author{Seonwook Park}
\affiliation{\institution{ETH Zurich}}
\email{spark@inf.ethz.ch}

\author{Xucong Zhang}
\affiliation{\institution{MPI for Informatics}}
\email{xczhang@mpi-inf.mpg.de}

\author{Andreas Bulling}
\affiliation{\institution{MPI for Informatics}}
\email{bulling@mpi-inf.mpg.de}

\author{Otmar Hilliges}
\affiliation{\institution{ETH Zurich}}
\email{otmarh@inf.ethz.ch}

\renewcommand{\shortauthors}{Park et al.}

\begin{document}
\title[Learning to Find Eye Region Landmarks for Remote Gaze Estimation]{Learning to Find Eye Region Landmarks for Remote Gaze Estimation in Unconstrained Settings}

\newcommand{\wookie}[1]{\noindent\textcolor{red}{\textbf{[Wookie:~}#1\textbf{]}}}
\newcommand{\xucong}[1]{\noindent\textcolor{blue}{\textbf{[Xucong:~}#1\textbf{]}}}
\newcommand{\andreas}[1]{\noindent\textcolor{orange}{\textbf{[Andreas:~}#1\textbf{]}}}
\newcommand{\otmar}[1]{\noindent\textcolor{ACMPurple}{\textbf{[Otmar:~}#1\textbf{]}}}

\newcommand{\etal}[0]{\textit{et~al.}}

\newcommand{\unity}[0]{\textsc{UnityEyes}}
\newcommand{\mpi}[0]{\textsc{MPIIGaze}}

\newcommand{\oh}[1]{\otmar{#1}}

\begin{abstract}

Conventional feature-based and model-based gaze estimation methods have proven to perform well in settings with controlled illumination and specialized cameras.
In unconstrained real-world settings, however, such methods are surpassed by recent appearance-based methods due to difficulties in modeling factors such as illumination changes and other visual artifacts.
We present a novel learning-based method for eye region landmark localization that enables conventional methods to be competitive to latest appearance-based methods.
Despite having been trained exclusively on synthetic data, our method exceeds the state of the art for iris localization and eye shape registration on real-world imagery.
We then use the detected landmarks as input to iterative model-fitting and lightweight learning-based gaze estimation methods.
Our approach outperforms existing model-fitting and appearance-based methods in the context of person-independent and personalized gaze estimation.

\end{abstract}



\maketitle

\section{Introduction}

Gaze estimation using off-the-shelf cameras, including those in mobile devices, can assist users with motor-disabilities or enable crowd-sourced visual saliency estimation without the cost of specialized hardware~\cite{xu2015turkergaze}.
Such methods can also improve user experience in everyday tasks, such as reading~\cite{biedert2010text,kunze13_iswc}, or facilitate gaze-based interaction~\cite{majaranta14_apc}.
However, existing gaze estimation systems can fail when encountering issues such as low image quality or challenging illumination conditions.
In this work, we provide a novel perspective on addressing the problem of gaze estimation from images taken in challenging real-world environments.

\begin{figure}
\raggedleft
\rotatebox{90}{\begin{minipage}[t][2mm]{11mm}\centering\scriptsize Columbia\end{minipage}}
\includegraphics[width=0.96\columnwidth]{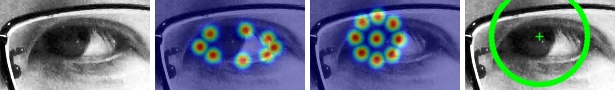}\\[0.3mm]
\rotatebox{90}{\begin{minipage}[t][2mm]{11mm}\centering\scriptsize UT\end{minipage}}
\includegraphics[width=0.96\columnwidth]{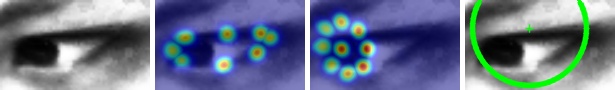}\\[0.3mm]
\rotatebox{90}{\begin{minipage}[t][2mm]{11mm}\centering\scriptsize MPIIGaze\end{minipage}}
\includegraphics[width=0.96\columnwidth]{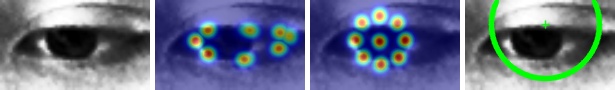}\\[0.3mm]
\rotatebox{90}{\begin{minipage}[t][2mm]{11mm}\centering\scriptsize EYEDIAP\end{minipage}}
\includegraphics[width=0.96\columnwidth]{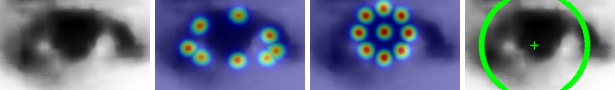}
\vspace*{-6mm}
\caption{
    Outputs from our landmark localization method on four datasets.
    From left to right: input eye image, eyelid landmarks, iris landmarks, eyeball center and radius estimates.
    Despite having been trained exclusively on synthetic data, our method applies directly to real-world eye images.
}
\label{fig:teaser}
\vspace*{-4.0mm}
\end{figure}

Conventional feature-based and model-based gaze estimation typically rely on accurate detection of eye region landmarks, such as the iris center or the eye corners.
Many previous works have therefore focused on accurately localizing iris center and eye corners landmarks \cite{Li2005CVPR,Timm2011VISAPP,Valenti2012TIP,Wood2014ETRAEA,Fuhl2016Ubicomp}.
On the recent real-world MPIIGaze dataset \cite{Zhang2015CVPR}, however, appearance-based methods were shown to significantly outperform model or feature-based methods such as EyeTab, a state-of-the-art model-based method \cite{Wood2014ETRAEA}.
The latest appearance-based methods perform particularly well in the person-independent gaze estimation task and in unconstrained settings in which visual artifacts, such as motion blur or sensor noise, are prevalent and cannot be easily removed or modeled \cite{Zhang2015CVPR,Zhang2017CVPRW,Krafka2016CVPR}. However, appearance based methods are not without drawbacks. First, training data is notoriously expensive and tedious to acquire and even if data is available the quality of the labels can vary. Furthermore, while effective most appearance-based methods are black-box solutions and understanding why and when they work can be a challenge. Finally, deep CNN architectures require a lot of computational power during training and adaptation once fully converged can be difficult and computationally costly.

In this work, we propose a new approach to gaze estimation that brings together the best of two-worlds. Similar to appearance-based methods, we leverage deep neural networks for representation learning. While prior work \emph{implicitly} learns to extract features that are useful for the gaze estimation task, we \emph{explicitly} learn features that are \emph{interpretable}. These interpretable features then allow traditional model-based or feature-based approaches to out-perform appearance-based methods in cross-dataset and person-specific gaze estimation.
In prior work, features were hand-crafted for gaze estimation using image processing techniques and model fitting. Since such approaches make assumptions about the geometry and shape of eyes, they are sensitive to appearance changes that are prevalent in unconstrained real-world images. Thus such methods suffer from the lack of robust detection of important features in such natural images.
We note that the task of eye-region landmark detection bears similarities to the problem of joint detection in human hand and full body pose. Thus, we overcome the above limitation by showing that robust eye-region landmark detectors can be trained \emph{solely} on high-quality synthetic eye images, providing detailed and accurate labels for the location of important landmarks in the eye region, such as eyelid-sclera border, limbus regions (iris-sclera border), and eye corners.
We then show that a relatively compact (in terms of model complexity) state-of-the-art convolutional neural network (CNN) can be trained on such synthetic data to robustly and accurately estimate eye region landmarks in \emph{real-world} images, without ever providing such images at training time (see Fig.~\ref{fig:teaser}).
The key advantage of this approach is that model-based and feature-based gaze estimation methods can be applied even to eye images for which iris localization and ellipse fitting can be highly challenging with traditional methods.
We experimentally show that such methods perform much better with learned eye landmarks than what has been reported in the literature previously.
This allows us to combine the well-studied task of landmark localization with personalized models or parameters for accurate eye tracking for individuals in the real-world.

In summary, the key contributions in our work are:
\begin{inparaenum}[(a)]
\item learning of a robust and accurate landmark detector on synthetic images only,
\item improvements to iris localisation and eyelid registration tasks on single eye images in real-world settings,
and
\item increased gaze estimation accuracies in cross-dataset evaluations as well as with as few as $10$ calibration samples for person-specific cases.
\end{inparaenum}

\begin{figure*}[h]
	\includegraphics[width=\textwidth]{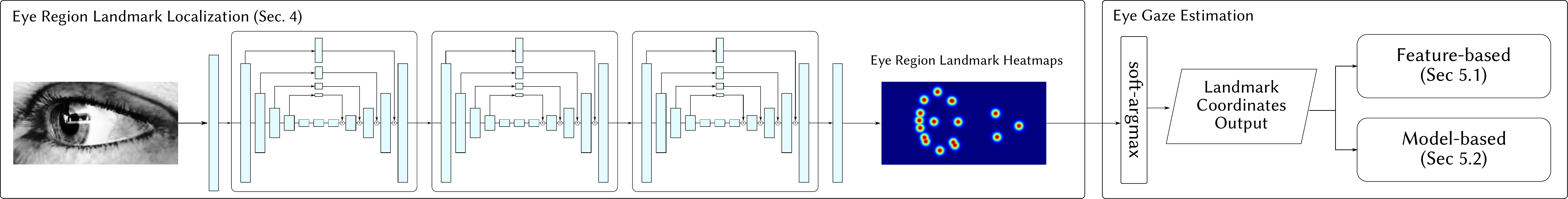}
	\vspace*{-6mm}
	\caption{
		Our architecture estimates eye region landmarks with a stacked-hourglass network trained on synthetic data (UnityEyes), evaluating directly on real, unconstrained eye images. The landmark coordinates can be used directly for model or feature-based gaze estimation.
	}
	\label{fig:hourglass}
	\vspace*{-3mm}
\end{figure*}

\section{Related Work}

Our method connects to a wide range of work in gaze estimation.
We provide a brief overview in this section specifically in the context of gaze estimation using off-the-shelf cameras.

\subsection{Feature-based Gaze Estimation}
Feature-based gaze estimation uses geometric considerations to hand-craft feature vectors which map the shape of an eye and other auxiliary information, such as head pose, to estimate gaze direction.
\citet{Huang2014MM} formulate a feature vector from estimated head pose and distance between 6 landmarks detected on a single eye.

A simple and common approach, as introduced by \citet{Sesma2012ETRA}, is the pupil-center-eye-corner vector or PC-EC vector that has later been adapted and successfully used for estimating horizontal gaze direction on public displays~\cite{Zhang14_AVIb,zhang13_chi}.
The specific claim of Sesma~\etal~\shortcite{Sesma2012ETRA} is that the PC-EC vector can replace the traditional corneal reflections that are traditionally used in eye tracking but that cannot be determined without IR illumination.
In addition, methods have been proposed that use Laplacian of Gaussian \cite{Huang2017MVA}, Histogram of Gaussian features \cite{FunesMora2016IJCV,Huang2017MVA} or Local Binary Patterns \cite{Huang2017MVA} among other features.

We show that using rich eye region landmarks detected by our model, when combined with features such as the PC-EC vector, significantly improves gaze estimation accuracy.

\subsection{Model-based Gaze Estimation}
The eyeball can generally be regarded as two intersecting spheres with deformations.
This is exploited in 3D model-based gaze estimation methods in which the center and radius of the eyeball as well as the angular offset between visual and optical axes are determined during user calibration procedures \cite{Xiong2014Ubicomp,Sun2015Elsevier,Wood2016ECCV,Wang2017ICCV}.
The eyeball center can be determined relative to a facial landmark (such as tip of nose) \cite{Xiong2014Ubicomp} or by fitting deformable eye region models \cite{Wood2016ECCV,Wang2017ICCV}.
In contrast, 2D model-based methods can observe the deformation of the circular iris due to perspective \cite{Wang2003ICCV,Wood2014ETRAEA}.

While previous works rely on accurate models of the face or eye region, our approach uses a neural network to fit an eyeball to an eye image.
This simple approach out-performs all prior works.

\subsection{Cross-ratio based Gaze Estimation}
In contrast to feature-based and model-based methods, cross-ratio methods utilise just a few IR illumination sources and the detection of their corneal reflections to achieve gaze estimation robust to head pose changes \cite{Yoo02}.
Recent extensions have improved this approach further via learning from simulation \cite{Huang14ETRA} and using multiple viewpoints \cite{Arar17arXiv}.
While these methods are promising, additional illumination sources may not be available on unmodified devices or settings for applications such as crowd-sourced saliency estimation using commodity devices.

\subsection{Appearance-based Gaze Estimation}
While appearance-based gaze estimation is a well-established research area \cite{Tan02WACV,Baluja1994}, it has only recently become possible to benchmark gaze estimation methods for in-the-wild settings with MPIIGaze \cite{Zhang2015CVPR} and GazeCapture \cite{Krafka2016CVPR} datasets.
Unlike early works which directly use image intensities as features to variants of linear regression \cite{Lu2011BMVC, Lu2011ICCV, FunesMora2014ETRA}, random forests \cite{Sugano2014CVPR}, and k-NN \cite{Wood2016ETRA}, recent works employ complex models such as CNNs \cite{Zhang2015CVPR,Krafka2016CVPR,Zhang2017CVPRW,Zhang2017PAMI} or GANs \cite{Shrivastava2017CVPR}.

Of particular interest are CNN-based approaches that have been demonstrated to yield high accuracy and have benefited from adopting novel architectures such as shown in \cite{Zhang2017PAMI} where a VGG-16 network yielded an improvement of $0.8^\circ$ ($6.3^\circ\rightarrow 5.5^\circ$) compared to the MnistNet architecture for within-dataset leave-one-person-out evaluation.
On the architectural side, other works investigated multi-modal training, such as with head pose information \cite{Zhang2015CVPR}, full-face images \cite{Zhang2017CVPRW}, or an additional ``face-grid'' modality for direct estimation of point of regard \cite{Krafka2016CVPR}.
The use of a lightly modified AlexNet with face images in \cite{Zhang2017CVPRW} and the drastic improvement in accuracy ($6.7^\circ\rightarrow 4.8^\circ$, within-dataset) highlight the need to further study what a CNN learns for the task of gaze estimation.

For the first time, we show that features implicitly learned by CNNs can be used for personalized gaze estimation using as few as $10$ calibration samples.
We also show that our explicitly learned landmarks features out-perform AlexNet features in this setting.

\subsection{Human Pose Estimation and Facial Landmark Localization}
The detection of so-called 2D landmarks from images is a well-studied topic in computer vision.
In particular, there exists a wealth of research on facial landmark detection and skeletal joint detection for the task of human pose estimation using deep convolutional neural networks \cite{Toshev2014CVPR,Sun2013CVPR}.
Facial landmarks and human joint positions can often be occluded but long-range dependencies and global context (i.e., other body parts) can make up for lack of local appearance based information.
Thus, recent work has attempted to learn spatial relations between joint positions \cite{Tompson2014NIPS}.
Of particular note is the stacked hourglass architecture by \citet{Newell2016ECCV}.
The stacked multi-scale architecture is simple and has been shown to out-perform other state-of-the-art methods while having low model complexity (few number of parameters).
Originally developed for pose estimation, the architecture has been successfully adapted to the task of facial landmark localization in the new Menpo Facial Landmark Localisation Challenge \cite{Zafeiriou2017CVPRW,Yang2017CVPRW}.

\section{Overview}
\label{sec:overview}

The goal of our work is to provide a robust and accurate landmark detector for eye-gaze estimation and related tasks. For this purpose we leverage a high quality synthetic eye image dataset \cite{Wood2016ETRA}. Based on recent progress in human pose estimation, we show how this data can be used to train such a detector and apply it directly to real images without fine-tuning or domain adaptation.
The extracted landmarks can be directly used in feature-based and model-based gaze estimation yielding improved accuracy in the cross-person and person-specific cases.
Thus, we bring such feature-based and model-based approaches back into the forefront of gaze estimation research in unconstrained settings.

At the core of our eye region landmark localization is a state-of-the-art CNN architecture, originally designed for the task of human-pose estimation.
The training data used is synthetic, and hence labels are correct even under heavy occlusion.
With appropriate training data augmentation, a robust model can be trained which detects equivalent landmarks on eye images captured in-the-wild
By using the rich landmarks-based features in simple learning-based methods such as SVR, we allow feature-based methods to perform comparably to appearance-based methods on webcam images.
In addition, the estimation of the eyeball center position and eyeball radius allows to fit a 3D eyeball model to any eye image, where camera intrinsic parameters are unknown.

In the next sections, we outline how eye region landmarks are detected in our approach (Sec.~\ref{sec:hourglass}), then describe two ways in which these landmarks can be used for gaze estimation: feature-based (Sec.~\ref{sec:feature-based}) and model-based (Sec.~\ref{sec:model-based}) methods.

\begin{figure}
	\includegraphics[width=\columnwidth]{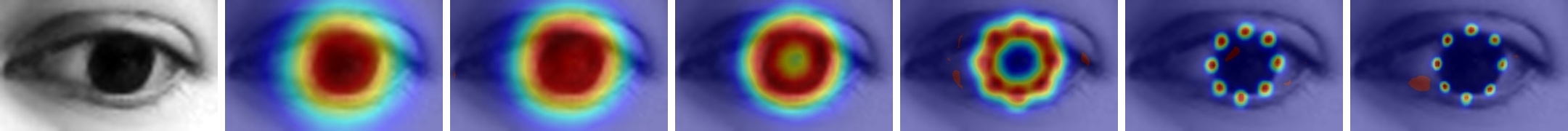}
	\includegraphics[width=\columnwidth]{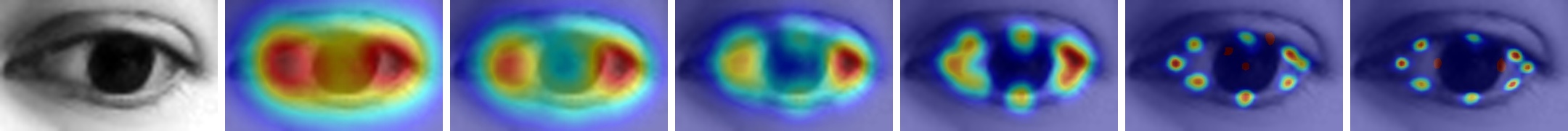}
	\vspace*{-6mm}
	\caption{
		As the model trains, its confidence in localizing specific iris (above) and limbus region (below) landmarks increases.
		Shown is an example from the MPIIGaze dataset.
	}
	\label{fig:heatmaps}
	\vspace*{-4mm}
\end{figure}

\section{Eye Region Landmark Localization}
\label{sec:hourglass}
In this section we describe the CNN architecture and training scheme used for eye region landmark localization.

\subsection{Architecture}
The hourglass network architecture~\cite{Newell2016ECCV} has previously been applied to human pose estimation, where a key problem is the occlusion of landmarks due to other body parts.
In such cases, the appearance of a landmark is no longer informative for accurate localization, and only prior knowledge can be used.
The hourglass architecture tries to capture long-range context by performing repeated improvement of proposed solutions at multiple scales, using so-called ``hourglass modules''.

Hourglass modules are similar to auto-encoders in that feature maps are downscaled via pooling operations, then upscaled using bilinear interpolation.
At every scale level, a residual is calculated and applied via a skip connection from the corresponding layer on the other side of the hourglass.
Thus when given $64$ feature maps, the network refines them at $4$ different image scales, multiple times.
This repeated bottom-up, top-down inference ensures a large effective receptive field and allows for the encoding of spatial relations between landmarks, even under occlusion.

In our work we adapt the original architecture to the task of landmark detection in eye-images (see Fig. \ref{fig:hourglass}). While eye images contain fewer global structuring elements than in pose estimation, there is still significant spatial context that can be exploited by-large receptive field models.
We take advantage of this property to detect eyeball center and occluded iris edge landmarks to reasonable accuracy, sometimes even under total occlusion.

The 64 refined feature maps can be combined via a $1\times 1$ convolutional layer to produce $18$ heatmaps (or confidence maps), each representing the estimated location of a particular eye region landmark.
Intermediate supervision is carried out by calculating a pixel-wise sum of squared differences loss on each predicted heatmap.
The original paper~\cite{Newell2016ECCV} demonstrates that using 8 hourglass modules with intermediate supervision yields significantly improved landmark localization accuracy compared to 2-stack or 4-stack models with the same number of model parameters.

In our work we use only 3 hourglass modules (with 1 residual module per stage), training on eye images and annotations provided by UnityEyes.
Though this model consists of less than 1 million model parameters, it is sufficient to demonstrate our approach and allows for a real-time implementation ($\sim20Hz$).
We use single eye images ($150\times 90$) as input and generate $18$ heatmaps ($75\times 45$): $8$ in the limbus region, $8$ on the iris edge, $1$ at the iris center, and $1$ at the eyeball center.
Fig.~\ref{fig:hourglass} shows our pipeline at inference time, where the network produces $18$ heatmaps which are further processed via a soft-argmax layer \cite{Honari2018CVPR} to find sub-pixel landmark coordinates.
These coordinates are then passed on to gaze estimation methods detailed in the following sections.

We find that by applying training data augmentation, a robust model can be learned even when training purely on synthetic eye images.
UnityEyes is effectively infinite in size and was designed to exhibit good variations in iris colour, eye region shape, head pose and illumination conditions.
Though the appearance variations do not include visual artifacts common in webcam images nor common eye decorations such as eyeglasses or make-up, we later show that our model applies directly to real-world imagery in the tasks of iris localization, eyelid registration, and gaze estimation.
The application of our model to an image from MPIIGaze is shown in Fig.~\ref{fig:heatmaps} where it can be seen that heatmaps become more accurate and confident as training progresses.

\subsection{Learning}
We now detail the training and data augmentation procedures.

\subsubsection{Loss Functions}
The network performs the task of predicting heatmaps, one per eye region landmark.
The heatmaps encode the per-pixel confidence on a specific landmark's location.
As such, the highest confidence value is at the pixel nearest to the actual landmark, with confidence quickly dropping off with distance and most of the map containing values set to $0$.
We place 2-dimensional Gaussians centered at the sub-pixel landmark positions such that the peak value is $1$.
The neural network then minimizes the $l_2$ distance between the predicted and ground-truth heatmaps per landmark via the following loss term:
\begin{equation}
	\mathcal{L}_{heatmaps} = \alpha \sum_{i=1}^{18} \sum_\mathbf{p} \left|\left| \tilde{h}_i(\mathbf{p}) - h_i(\mathbf{p}) \right|\right|^2_2 \quad ,
\end{equation}
where $h(\mathbf{p})$ is the confidence at pixel $\mathbf{p}$ and $\tilde{h}$ is a heatmap predicted by the network.
We empirically set the weight coefficient $\alpha = 1$.

For our model-based method, we additionally predict an eyeball radius value $\tilde{r}_{uv}$.
This is done by first appending a soft-argmax layer \cite{Honari2018CVPR} to calculate landmark coordinates from heatmaps, then further appending $3$ linear fully-connected layers with $100$ neurons each (with batch normalization \cite{Ioffe15ICML} and ReLU activation) and one final regression layer with $1$ neuron.
The loss term for the eyeball radius output is:
\begin{equation}
	\mathcal{L}_{radius} = \beta \left|\left| \tilde{r}_{uv} - r_{uv} \right|\right|^2_2 \quad ,
\end{equation}
where we set $\beta = 10^{-7}$ and use ground-truth radius $r_{uv}$.

\subsubsection{Training Data Augmentation}

\begin{figure}
\begin{subfigure}[t]{0.48\columnwidth}
	\centering
	\includegraphics[width=0.3\columnwidth]{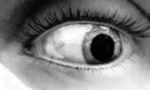}
	\includegraphics[width=0.3\columnwidth]{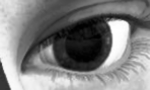}
	\includegraphics[width=0.3\columnwidth]{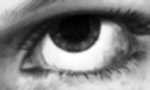} \\[-1pt]
	\includegraphics[width=0.3\columnwidth]{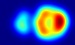}
	\includegraphics[width=0.3\columnwidth]{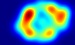}
	\includegraphics[width=0.3\columnwidth]{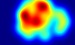}
	\caption{Min. data augmentation}
\end{subfigure}
\begin{subfigure}[t]{0.48\columnwidth}
	\centering
	\includegraphics[width=0.3\columnwidth]{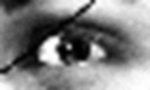}
	\includegraphics[width=0.3\columnwidth]{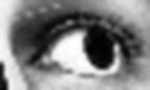}
	\includegraphics[width=0.3\columnwidth]{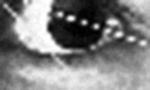} \\[-1pt]
	\includegraphics[width=0.3\columnwidth]{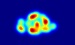}
	\includegraphics[width=0.3\columnwidth]{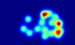}
	\includegraphics[width=0.3\columnwidth]{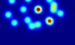}
	\caption{Max. data augmentation}
\end{subfigure}
\begin{subfigure}[t]{0.94\columnwidth}
  \centering
  \vspace*{2mm}
  \includegraphics[width=\columnwidth]{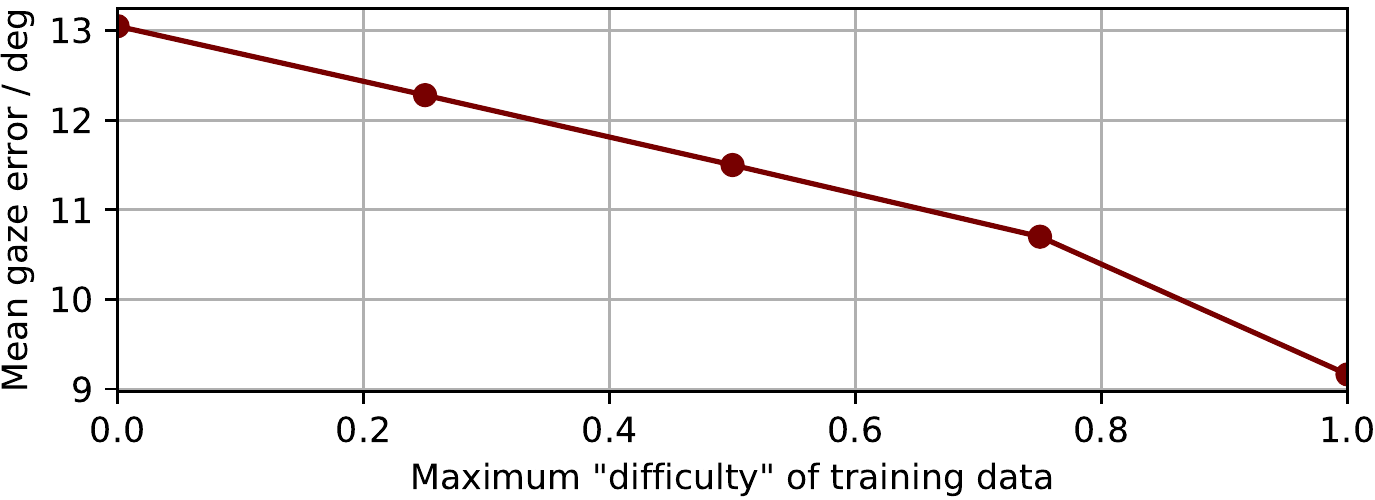}
  \vspace*{-5mm}
  \caption{Model-based gaze estimation error on MPIIGaze.}
  \vspace*{-2mm}
  \label{fig:data_aug_graph}
\end{subfigure}
\caption{Example UnityEyes input images with minimum (a) and maximum (b) data augmentation.
The 18 individual heatmaps are combined into one for visualization purposes.
(c) shows that increased data augmentation yields higher gaze estimation accuracy on real-world images.
}
\label{fig:data_aug_images}
\vspace*{-4mm}
\end{figure}

We found that employing strong data augmentation during training improves the performance of the model in the context of gaze estimation.
We apply the following augmentations (range in brackets are scaling coefficients of value sampled from $\mathcal{N}(0,1)$): translation (2--10 px),
rotation (0.1--2.0 rad),
intensity (0.5--20.0),
blur (0.1--1.0 std. dev. on $7\times 7$ Gaussian filter),
scale (1.01--1.1),
downscale-then-upscale (1x--5x),
and addition of lines (0--2) for artificial occlusions.
We do not perform any image flipping during training but simply assure that the inner eye corner is on the left side of the input image at test time.

Our final training scheme applies curriculum learning, increasing noise as training progresses \cite{Bengio2009ICML}.
To make this easier to control, we implement a \emph{difficulty}-measure which ranges from $0$ to $1$.
We begin training with difficulty $0$ and linearly increase difficulty until $10^6$ training steps have passed.
Thereafter, difficulty is kept at $1$.
Sample input images are shown in Fig.~\ref{fig:data_aug_images}.

We further verify the utility of strong and extensive data augmentation by performing cross-dataset gaze estimation on MPIIGaze with manual eye corner annotations, using our model-fitting method (see Sec. \ref{sec:model-based}).
Fig.~\ref{fig:data_aug_graph} shows a significant decrease in gaze estimation error with higher amounts of training data augmentation (after 1M training steps with batch size of 32).
We also observe that with more training steps, a model trained with weaker data augmentation can perform similarly to that trained with stronger augmentation.
Thus data augmentation not only improves the robustness of the model but speeds up training.

\subsubsection{Further Details}
We use the ADAM optimizer \cite{Kingma2014arXiv}, with a learning rate of $5\times10^{-4}$, batch size of $16$, $l_2$-regularization coefficient of $10^{-4}$, and ReLU activation.
Our reference model was trained for $6.8$M steps on an Nvidia 1080 Ti GPU.
During test time, we use statistics computed per-batch for batch normalization, instead of using the population statistics computed during training, from synthetic data.

\section{Gaze Estimation}
In this section we discuss how we make use of our eye region landmarks in feature-based and model-based gaze estimation.

\subsection{Feature-based gaze estimation}
\label{sec:feature-based}
To create our features, we first consider the inner and outer eye corners, $\mathbf{c}_1$ and $\mathbf{c}_2$ respectively.
We normalize all detected landmark coordinates by the eye width $\mathbf{c}_2 - \mathbf{c}_1$, and center the coordinate system on $\mathbf{c}_1$.
In addition, we provide a 2D gaze prior by subtracting eyeball center $(u_c,\,v_c)$ from iris center $(u_{i0},\,v_{i0})$.
Despite being a crude estimate, this prior improves performance significantly where very low number of training samples are available (such as in person-specific calibration).
Our final feature vector is formed of 17 normalized coordinates (8 from limbus, 8 from iris edge, 1 from iris center) and a 2D gaze direction prior, resulting in 36 features.

The 36 landmarks-based features are then used to train a support vector regressor (SVR) which directly estimates a gaze direction in 3D, $(\theta, \phi)$ representing eyeball pitch and yaw respectively.
The SVR can be trained to be person-independent with a large number of images from different people, or from a small set of person-specific images for personalized gaze estimation.
Where possible, we perform leave-one-out cross validation to determine hyper parameters.

\subsection{Model-based gaze estimation}
\label{sec:model-based}
\begin{figure}
	\includegraphics[width=0.8\columnwidth]{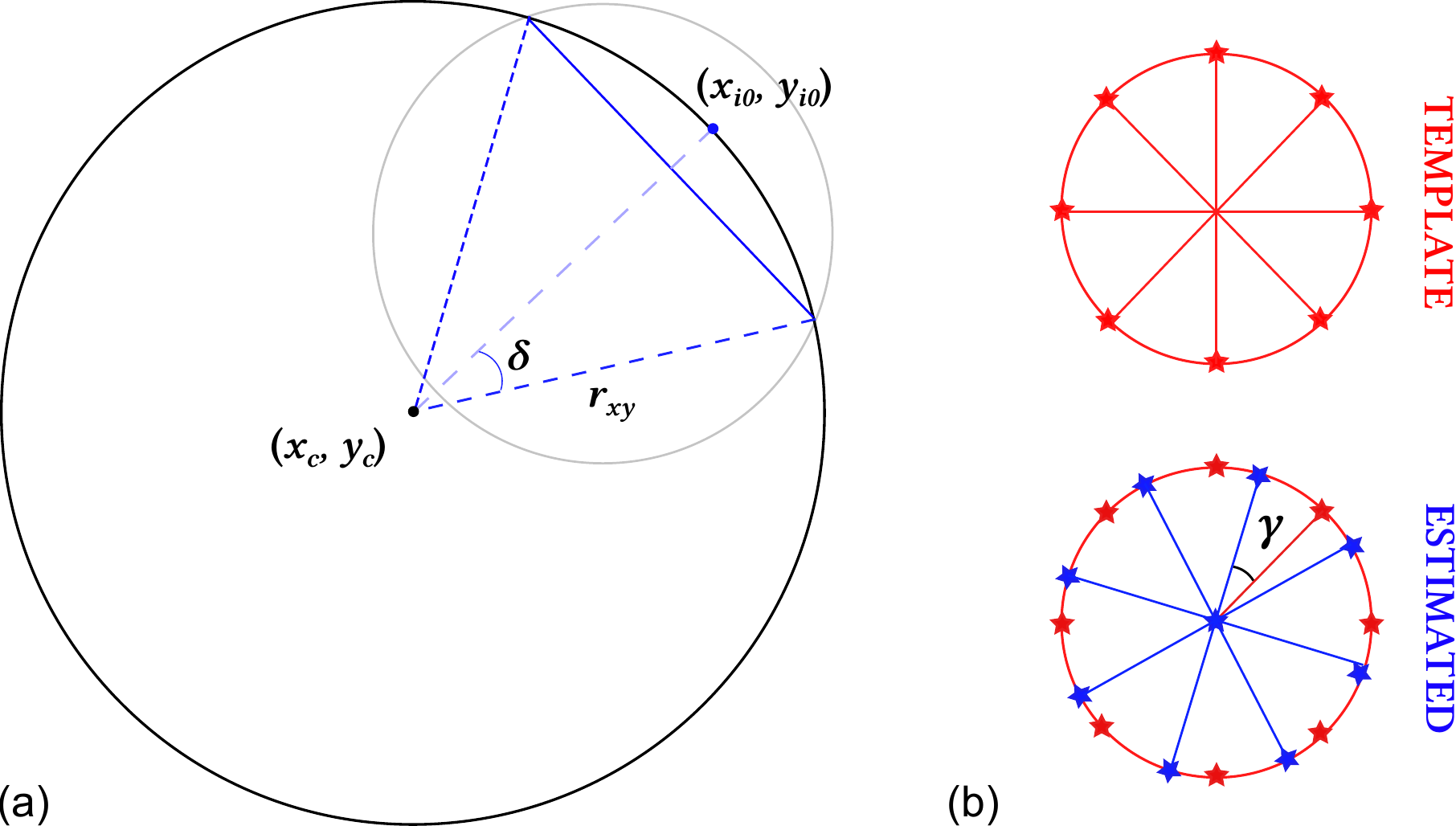}
	\caption{The eyeball is modeled as two intersecting spheres. Our iterative fitting is performed by predicting and matching the 8 iris edge landmarks and iris center landmark.}
	\label{fig:eyeball}
	\vspace*{-5mm}
\end{figure}
As done commonly for remote gaze estimation \cite{Sun2015Elsevier,Wood2015ICCV}, we use a simple model of the human eyeball, depicting it as one large sphere and a smaller intersecting sphere to represent the corneal bulge (Fig.~\ref{fig:eyeball}a).
Let us denote the predicted 8 iris edge landmarks in a given eye image as $\left(u_{i1},v_{i1}\right),\,\ldots\,,\left(u_{i8},\,v_{i8}\right)$.
In addition, we detect 1 landmark for the eyeball center $\left(u_c,\,v_c\right)$ and another for the iris center $\left(u_{i0},\,v_{i0}\right)$.
Furthermore we estimate the eyeball radius in pixels, $r_{uv}$.
Knowing the eyeball and iris center coordinates and eyeball radius in pixels makes it possible to fit a 3D model without access to any camera intrinsic parameters, and thus without the need for camera calibration.

As we assume no known camera parameters, our coordinates can only be unprojected into 3D space in pixel units.
Thus, the radius remains $r_{xy} = r_{uv}$ in 3D space and $(x_c,\,y_c)=(u_c,\,v_c)$.
If we assume a gaze direction $(\theta,\,\phi)$, we can now write the iris center coordinates as:
\begin{equation}
\begin{split}
	u_{i0} &= x_{i0} = x_c - r_{xy} \cos\theta \sin\phi \\
	v_{i0} &= y_{i0} = y_c + r_{xy} \sin\theta
\end{split}
\end{equation}

\noindent
To write similar expressions for the 8 iris edge landmarks, we must jointly estimate angular iris radius $\delta$ and an angular offset $\gamma$ which is equivalent to eye roll.
This offset between template and actual eye roll or iris rotation is shown in Fig.~\ref{fig:eyeball}b.
With the new variables we can now write for the $j$-th iris edge landmark (with $j = 1\ldots 8$):
\begin{equation}
\begin{split}
	u_{ij} &= x_{ij} = x_c - r_{xy} \cos\theta^\prime_j \sin\phi^\prime_j \\
	v_{ij} &= y_{ij} = y_c + r_{xy} \sin\theta^\prime_j
\end{split}
\end{equation}
where,
\begin{equation}
\begin{split}
	\theta^\prime_j &= \theta + \delta \sin\left(\frac{\pi}{4}j + \gamma\right) \\
	\phi^\prime_j   &= \phi + \delta \cos\left(\frac{\pi}{4}j + \gamma\right) \\
\end{split}
\end{equation}

\noindent
For model-based gaze estimation, $\theta$, $\phi$, $\gamma$, and $\delta$ are unknown whereas other variables are provided by the eye region landmark localization step of our system.
We solve this problem using an iterative optimization method such as the conjugate gradient method where the minimized loss function is represented as:
\begin{equation}
	\sum_{0\leq j\leq 8}\left(u_{ij} - u^\prime_{ij}\right)^2 +
	\left(v_{ij} - v^\prime_{ij}\right)^2
\end{equation}

\noindent
where $\left(u^\prime_{ij},\,v^\prime_{ij}\right)$ is the estimated pixel coordinates of the $j$-th iris landmark at each iteration.

To adapt this model to a specific person, we calculate person-specific parameters based on calibration samples.
Gaze correction can be applied with
$\left(\tilde{\theta},\,\tilde{\phi}\right) = \left(\theta+\Delta\tilde{\theta},\,\phi+\Delta\tilde{\phi}\right)$ where
$\left(\Delta\tilde{\theta},\,\Delta\tilde{\phi}\right)$ is the person-specific angular offset between optical and visual axes.

\section{Evaluations}
We now provide comprehensive evaluations of our method.
Since our method can output many more eye landmarks, there is no directly comparable baselines from previous works.
Therefore, we first evaluate how well our method can estimate eye-shape by assessing eyelid registration performance, and then address the problem of iris center localization on remote eye images.
Finally, we perform various gaze estimation evaluations where we:
\begin{inparaenum}[(a)]
\item evaluate the accuracy of our model-fitting approach,
\item evaluate cross-dataset performance of our model-based and feature-based methods against an appearance-based method,
and
\item compare feature-based, model-based, and appearance-based approaches for the case of training a gaze estimation model on few calibration samples.
\end{inparaenum}

For our evaluations of gaze estimation error, we select the EYEDIAP \cite{FunesMora2014ETRA}, MPIIGaze \cite{Zhang2015CVPR}, UT Multiview \cite{Sugano2014CVPR} and Columbia Gaze \cite{Smith2013UIST} datasets.
These datasets are often used to evaluate gaze estimation methods from remote camera imagery.
For the EYEDIAP dataset, we evaluate on VGA images with static head pose for fair comparison with similar evaluations from \cite{Wang2017ICCV}.
Please note that we do not discard the challenging floating target sequences from EYEDIAP.
For MPIIGaze, we use manually annotated eye corners to produce segmented eye images to ensure that we can detect eye region landmarks within image bounds.
The input eye image dimensions to the hourglass network are $150\times 90$, unless stated otherwise.
Specific challenges represented in the selected datasets include far distance from camera to eye (EYEDIAP), low image quality (MPIIGaze, UT Multiview), and high variations in head poses (UT Multiview) or gaze directions (EYEDIAP).

\subsection{Eyelid Registration}

\begin{figure}[h]
	\vspace*{-2mm}
	\centering
	\includegraphics[width=1.0\columnwidth]{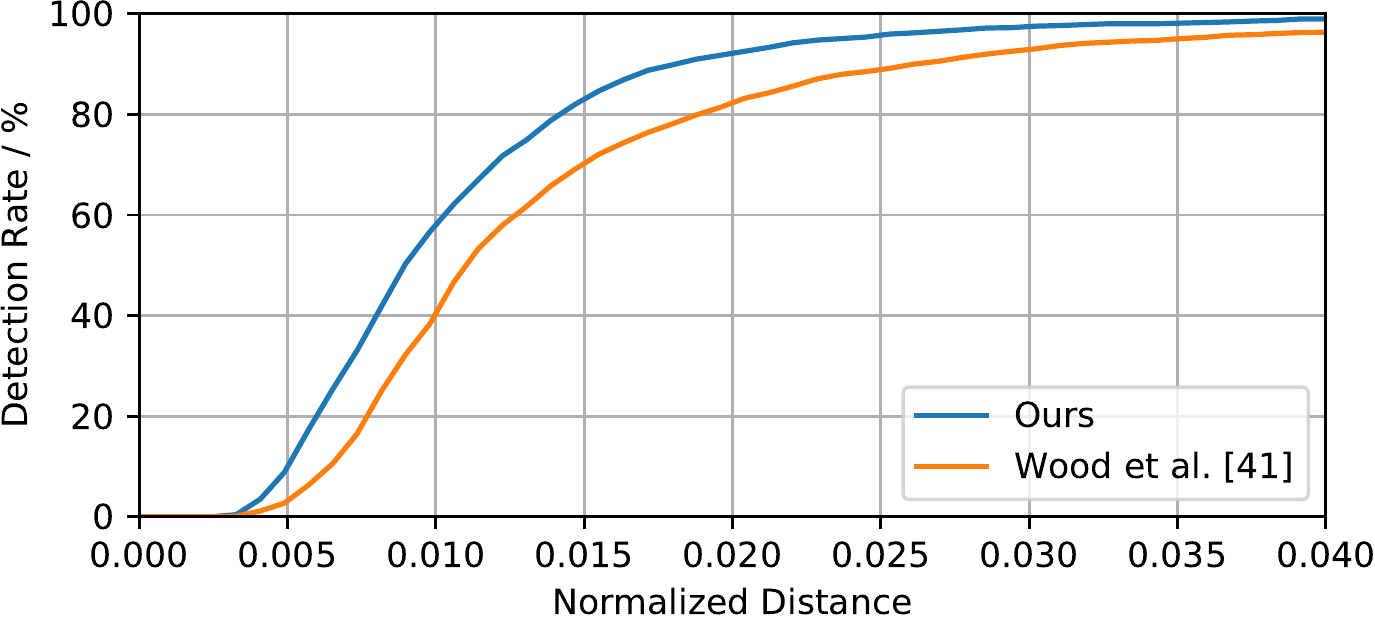}
	\vspace*{-6.5mm}
	\caption{Eyelid registration performance compared against \citet{Wood2015ICCV}. Success rate is thresholded on euclidean distance error normalized by interocular distance.
	}
	\label{fig:eyelid_reg}
	\vspace*{-5mm}
\end{figure}
The method proposed by \citet{Wood2015ICCV} is the current state-of-the-art for eyelid registration on challenging remote eye images collected in real-world environments.
We first perform eyelid registration on the subset of the 300-W dataset~\cite{Sagonas2013ICCVW} as was tested in \cite{Wood2015ICCV}.
The eyelid registration error for a single eye is defined as the mean Euclidean distance to the annotated eyelid, normalized by interocular distance.
The ground-truth eyelid annotation is generated from the 6 annotated eyelid landmarks and
interocular distance is approximated by the distance between the outer corners of the left and right eyes.
Fig.~\ref{fig:eyelid_reg} shows our results in comparison to the previous constrained local neural field (CLNF) approach where it can be seen that our method is more accurate and robust.
This clearly demonstrates the advantage of the proposed CNN-based eye landmark localization method.

\subsection{Iris Localization}

\begin{figure}[h]
\begin{subfigure}{0.34\columnwidth}
	\includegraphics[width=\columnwidth]{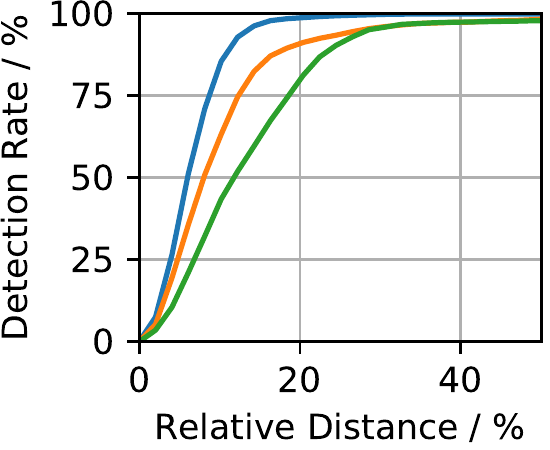}
	\caption{BioID}
\end{subfigure}
\hfill
\begin{subfigure}{0.31\columnwidth}
	\includegraphics[width=\columnwidth]{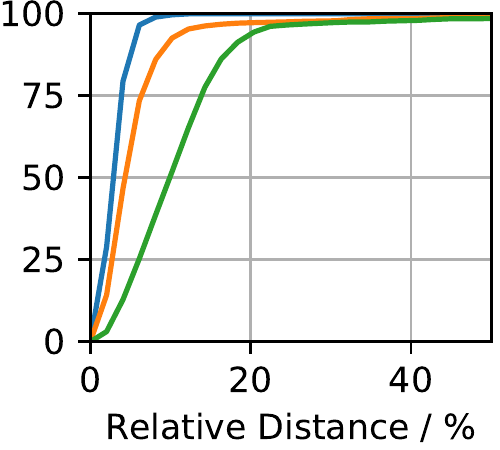}
	\caption{GI4E}
\end{subfigure}
\hfill
\begin{subfigure}{0.31\columnwidth}
	\includegraphics[width=\columnwidth]{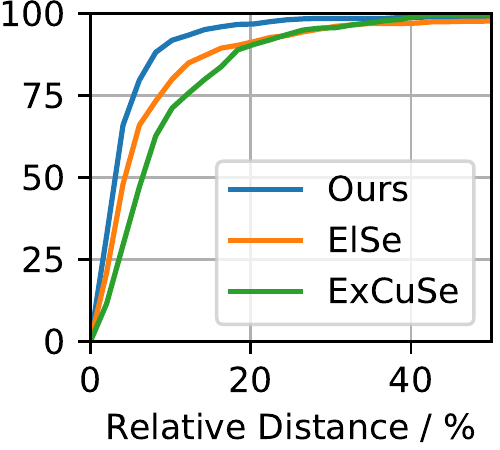}
	\caption{Fuhl~\etal}
\end{subfigure}
\vspace*{-3mm}
\caption{Iris localization success rate thresholded on euclidean distance error normalized by horizontal eye width. We compare our method with the state-of-the-art ExCuSe and ElSe algorithms which perform well on webcam images.}
\label{fig:iris}
\vspace*{-2mm}
\end{figure}

\citet{Fuhl2016Ubicomp} show that pupil detection algorithms developed for the head-mounted case can excel at iris center localization on eye images captured by remotely located cameras.
In such scenarios, no glint (from active illumination) can be perceived, and the pupil is often indistinguishable from iris regions.
ElSe \cite{Fuhl2016ETRA} is reported to perform best on BioID, GI4E \cite{Villanueva2013TOMM}, and a newly captured dataset exhibiting challenging head poses.

We compare our method, ElSe, and ExCuSe \cite{Fuhl2015CAIP} on the mentioned datasets and report our results in Fig.~\ref{fig:iris}.
Input eye images are created based on eye corner annotations such that the input resolution for our method is $150\times 90$ and for ExCuSe and ElSe are $384\times 288$ (as done in \cite{Fuhl2016Ubicomp}).
The figure reports iris localization success rates determined based on distance thresholds to ground-truth, normalized by horizontal eye width.
It can be seen that our method consistently out-performs the two state-of-the-art methods across the whole threshold range for all three datasets.

\subsection{Model-Based Gaze Estimation}

\begin{table}[h]
\vspace*{-2mm}
\caption{
	Mean angular error for our model-fit approach compared with state-of-the-art.
}
\label{tbl:model-based}
\vspace*{-2mm}
\centering
\begin{tabular}{l|ccccc|ccc}
\hline
& Columbia & EYEDIAP$^\star$ \\
\hline
\cite{Xiong2014Ubicomp} &         9.7 &          21.3 \\
\cite{Wood2016ECCV}     &         8.9 &          21.5 \\
\cite{Wang2017ICCV}     &         7.1 &          17.3 \\
Ours                    & \textbf{7.1} & \textbf{11.9} \\
\hline
\end{tabular}
\hspace*{11mm}
\begin{minipage}{0.9\columnwidth}
	\small
	\vspace{0.5mm}
	$^\star$~VGA images (V), static head pose (S).
\end{minipage}
\vspace*{-5mm}
\end{table}

Our model-fitting algorithm assumes no knowledge about camera intrinsic parameters or accurate 3D models of the face or eye region of specific persons.
It rather relies on an estimation of eyeball center, iris center landmarks, and eyeball radius.
We compare our model-fitting approach with the other state-of-the-art methods on both Columbia Gaze and EYEDIAP datasets, using $20$ images from each person for calibration.
For testing on EYEDIAP, we use all available frames to provide a comprehensive evaluation.
It can be seen from Tab.~\ref{tbl:model-based} that while our results are similar to \citet{Wang2017ICCV} for the Columbia Gaze dataset, we achieve significant improvements on EYEDIAP.
Considering that the VGA images from the EYEDIAP dataset are very low-resolution and of low quality, our results further demonstrate the robustness of our approach.
In contrast to previous model-fitting approaches which use full face or eye region images as input, we solve the more challenging problem of model-based gaze estimation from single eye images.
In this case eye rotation is ambiguous, in that it is not easily possible to define an ``up'' direction based on the eye shape of an unseen person.

\subsection{Feature-Based Gaze Estimation}

\begin{table}[h]
\caption{
	Mean gaze estimation error of different feature representations in SVR training when using $20$ calibration samples.
	\emph{Ours} refers to eyelid and iris landmarks as well as an iris-center-eyeball-center vector.
}
\label{tbl:features}
\vspace*{-1.5mm}
\centering
\begin{tabular}{>{\small}l|cccc}
	\hline
	Features & {\small EYEDIAP} & {\small MPIIGaze+} & {\small UT} & {\small Columbia} \\
	\hline
	Pupil Center          &         $8.4$  &         $5.3$  &         $17.2$  &         $8.0$  \\
	PC-EC vectors         &         $8.0$  &         $4.9$  &         $13.0$  &         $7.7$  \\
	Iris landmarks        &         $8.3$  &         $5.0$  &         $17.8$  &         $7.5$  \\
	Eyelid+Iris landmarks & $\mathbf{7.4}$ & $\mathbf{4.6}$ &         $12.4$  &         $6.5$  \\
	Ours                  &         $7.5$  & $\mathbf{4.6}$ & $\mathbf{11.5}$ & $\mathbf{6.2}$ \\
	\hline
\end{tabular}
\vspace*{-2mm}
\end{table}

In Sec.~\ref{sec:feature-based}, we argued for using all landmark coordinates and an iris-center-eyeball-center vector for feature-based gaze estimation.
Tab.~\ref{tbl:features} shows an evaluation of different input features when training an SVR with $20$ person-specific samples.
In each case, the coordinates or vectors are normalized by eye width, defined as the Euclidean distance between the detected eye corners.
It can be seen that na\"ive features, such as pupil center or iris landmark coordinates, do not result in good gaze estimation performance.
It is only when both eyelid and iris landmarks are used that feature-based gaze estimation can improve significantly, in particular for the UT Multiview dataset that contains large variability in head pose and gaze direction.
Performance is further improved by adding our gaze direction prior based on our eyeball center estimations.
For subsequent evaluations, we select eight iris edge landmarks, eight eyelid landmarks, one iris center landmark, and one iris-center-eyeball-center vector as features for training our feature-based method.

\subsection{Cross-Dataset Gaze Estimation}

\begin{table}[h]
\vspace*{-1.5mm}
\caption{
	Cross-dataset evaluation of gaze estimation error when trained on UT Multiview ($150$k entries).
}
\label{tbl:cross-dataset}
\vspace*{-1.5mm}
\centering
\begin{tabular}{l|ccc}
	\hline
	& EYEDIAP & MPIIGaze+ & Columbia \\
	\hline
	AlexNet                                  &         $37.1$  &         $12.5$ &         $12.0$ \\
	$SVR_\text{\scriptsize\emph{Landmarks}}$ & $\mathbf{23.3}$ &         $10.7$ &         $10.0$ \\
	model-fit                                &         $26.6$  & $\mathbf{8.3}$ & $\mathbf{8.7}$ \\
	\hline
\end{tabular}
\vspace*{-2mm}
\end{table}

For applications in public display settings \cite{zhang13_chi,Zhang14_AVIb,Sugano2016UIST}, it is particularly interesting to evaluate the person-independence of a proposed gaze estimation method.
We therefore evaluate our model-based and feature-based methods alongside an appearance-based method (AlexNet) for the case of training on the UT Multiview, and testing on EYEDIAP, MPIIGaze, and Columbia Gaze datasets.
Tab.~\ref{tbl:cross-dataset} shows that both of our proposed methods out-perform an AlexNet baseline.
Our evaluations on MPIIGaze+ in particular are competitive against the reported result of $9.8^\circ$ \cite{Zhang2017PAMI} using a VGG-16 based architecture.
Our model-fit approach achieves the lowest ever reported error of $8.3^\circ$ for the case of training on UT Multiview and testing on MPIIGaze+.

The task of landmark localization, when done accurately, allows for dataset-specific biases and artifacts to be abstracted away, leaving a pure representation of eye shape.
The findings here indicate that learning \emph{explicit}, interpretable features helps in significantly improving accuracy over the na\"ive appearance-based baseline which learns \emph{implicit} features which may or may not represent eye shape or gaze direction due to the lack of explicit constraints during training.
This is especially true for cross-dataset evaluations but also applies in general for person-independent gaze estimation.

Please also note the comparative model complexities of the approaches.
While AlexNet consists of over 80 million trainable model parameters, our network is formed of less than 1 million parameters.

\subsection{Personalized Gaze Estimation}

\begin{figure}[h]
	\begin{subfigure}{0.48\columnwidth}
		\includegraphics[width=\columnwidth]{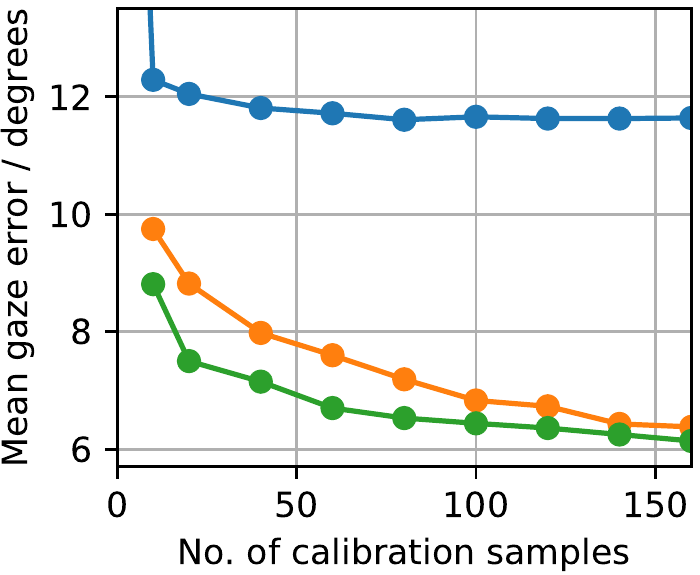}
		\vspace*{-5mm}
		\caption{EYEDIAP {\small(V, S)}}
		\vspace*{1mm}
	\end{subfigure}
	\hfill
	\begin{subfigure}{0.48\columnwidth}
		\includegraphics[width=\columnwidth]{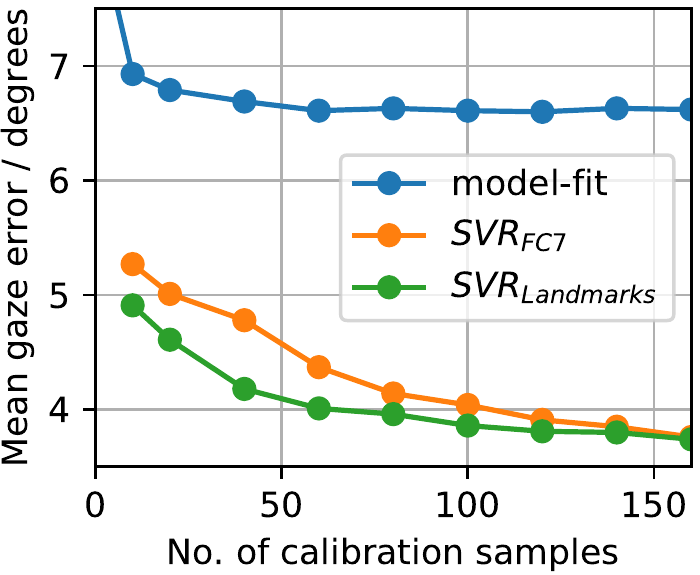}
		\vspace*{-5mm}
		\caption{MPIIGaze+}
		\vspace*{1mm}
	\end{subfigure}
	\begin{subfigure}{0.48\columnwidth}
		\includegraphics[width=\columnwidth]{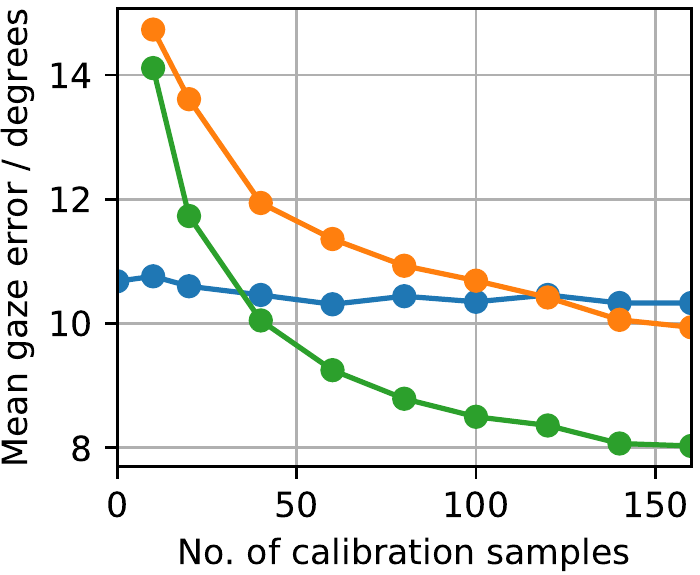}
		\caption{UT Multiview}
	\end{subfigure}
	\hfill
	\begin{subfigure}{0.48\columnwidth}
		\includegraphics[width=\columnwidth]{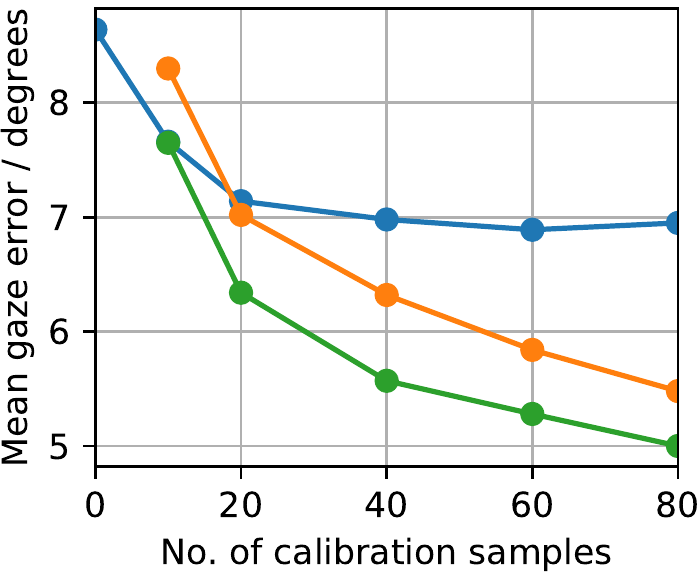}
		\caption{Columbia}
	\end{subfigure}
	\caption{
		Mean angular error for personalized gaze estimation where we compare our model-based and feature-based approaches against an appearance-based approach.
	}
	\label{fig:personalized}
	\vspace*{-2mm}
\end{figure}

While appearance-based gaze estimation has recently demonstrated progress towards person-independent gaze estimation in unconstrained settings, it is an open question whether an appearance-based approached can easily adapt to individual people (i.e., provide better than average performance on a specific user).

One advantage of our approach is that we introduce an eye region landmark localization method that does not require any calibration.
The detected landmarks can be directly used to train a simple regression method that in turn is amenable to personalization via \emph{few} calibration samples.
It has been demonstrated that such calibration samples can be collected using a conventional grid of 9-points and that this can be further boosted by recording short video clips \cite{Lu2011BMVC}.
In this evaluation, we explore personalization in this sense for our model-based and feature-based methods.

For a fair comparison to appearance-based methods, we train an AlexNet that directly regresses gaze direction using training data augmentation and UnityEyes images.
We then take the output of the 7th layer and train an SVR ($SVR_{FC7}$) on the $4,096$-dimensional feature vector. Thus we directly compare the utility of our \emph{explicit}, low-dimensional feature representation versus an \emph{implicit}, high-dimensional feature representation.
Note that this is a fair comparison since the last fully connected layer of the AlexNet directly sits under the regression layer during training, hence the learned representation is optimized for the task of gaze estimation.

For MPIIGaze, we select $3,000$ samples per person as done in \cite{Zhang2017PAMI} while for EYEDIAP and UT Multiview, we select $1,000$ samples per person.
We sample our calibration entries from these sets and use the remaining entries for evaluation.
For sampling calibration entries, we perform a variance-maximizing sampling based on ground-truth gaze pitch and yaw angles.

Our results in Fig.~\ref{fig:personalized} show that the model-fitting approach only benefits initially and only marginally from calibration.
We note that our calibration procedure for this case is simple and only estimates the offset between visual and optical axes.
However, for both $SVR_{FC7}$ and $SVR_{Landmarks}$, we see significant improvements in accuracy with increasing calibration samples.
When comparing $SVR_{Landmarks}$ and $SVR_{FC7}$, it can be seen that our landmarks-based method performs better across the whole range and accuracies only converge to similar values after more than $100$ calibration samples.
In particular in the case when there are only a \emph{low} number of calibration samples ours provides significant accuracy gains.
This suggests that personalized gaze estimation is indeed feasible with our feature-based method.
Furthermore, even with as few samples as $20$, our method improves significantly with 16.4\% over the state-of-the-art person-independent gaze estimation error as reported in \cite{Zhang2017PAMI} ($5.5^\circ \rightarrow 4.6^\circ$).
Finally, $SVR_{Landmarks}$ shows improvements of up to $2^\circ$ in terms of gaze estimation error compared to $SVR_{FC7}$ on EYEDIAP and UT Multiview which exhibit large variations in gaze direction and head pose respectively.

\subsection{Real-time Implementation}
\begin{figure}
	\includegraphics[width=\columnwidth]{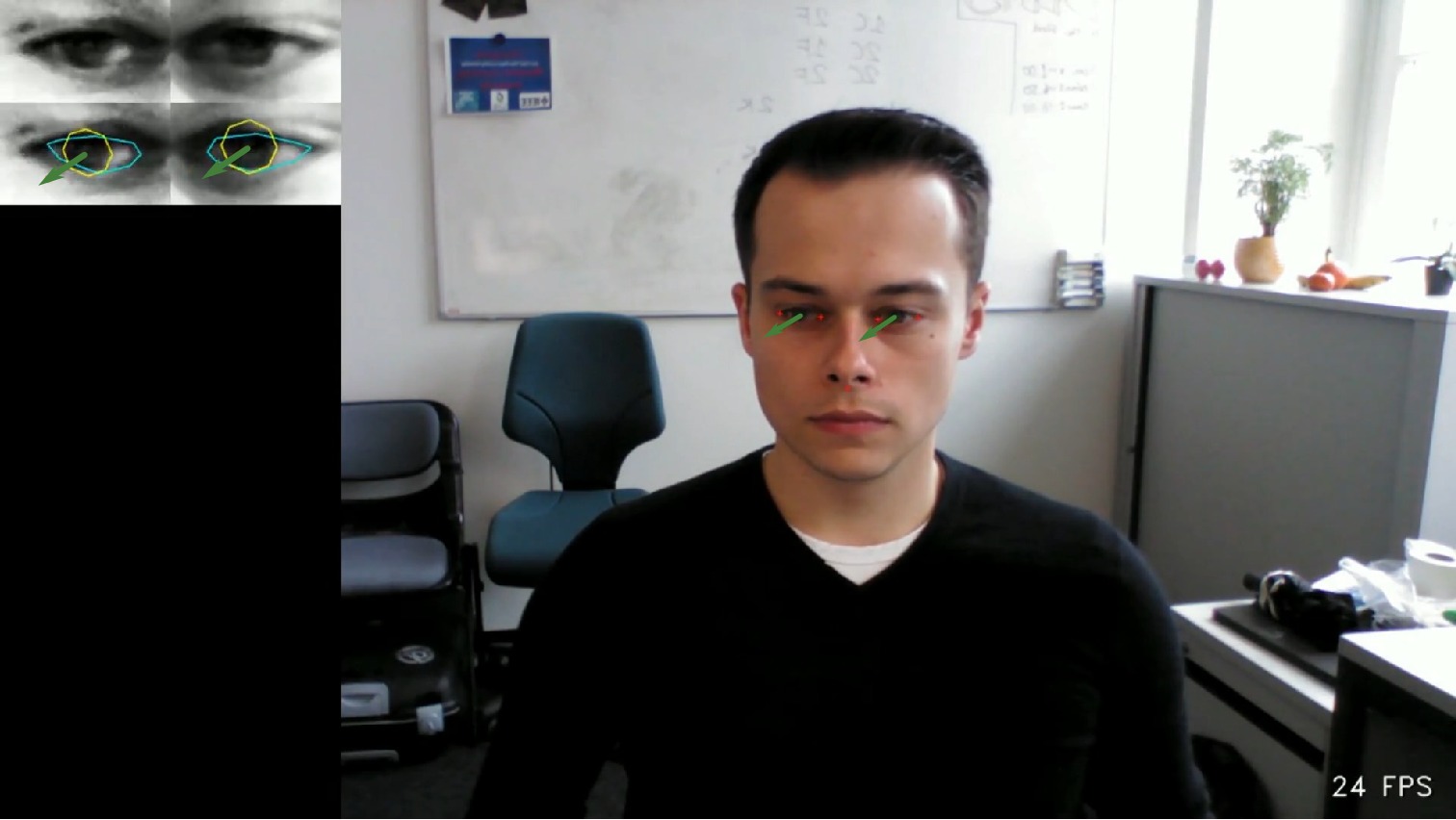}
	\vspace*{-7mm}
	\caption{
	Sample frame from our demo application.
	Displayed on the top-left are the segmented eye images used as input to our landmark localization.
	Gaze direction is estimated using a SVR and visualized as arrows (emphasized).
	}
	\label{fig:demo}
	\vspace*{-5mm}
\end{figure}
To provide an initial qualitative evaluation on how well our method works we implement a real-time proof-of-concept system.
The application begins by grabbing a $1280\times 720$ RGB frame from a Creative Senz3D webcam (where depth information is not used).
We use dlib \cite{dlib} for face detection and a $5$-point facial landmark detection.
Two eye images are segmented using the detected eye corners, then each are used as input to our eye landmark localization model.
The estimated iris and eyelid landmarks are visualized by blue and yellow outlines respectively in the top-left corner of Fig.~\ref{fig:demo}.
We then use a SVR trained on MPIIGaze and UT Multiview datasets to provide an estimation of gaze direction for each eye individually.
The unoptimized system runs at up to $26Hz$ on a desktop PC (Intel i7-4770 and Nvidia GeForce 1080Ti).
In the accompanying video\footnote{\url{https://youtu.be/cLUHKYfZN5s}}, we demonstrate robustness to challenging illumination conditions, occlusion due to eyeglasses, large head pose changes, as well as noise and blur due to the large distance between camera and person.

\section{Conclusions and Future Work}

\paragraph{Summary.}
In this paper we introduced a new approach to gaze estimation from webcam images in unconstrained settings.
We showed that eye region landmarks around the iris and eyelid edges can be found in real, unconstrained images using a model trained exclusively on synthetic input.
Furthermore, the same neural network can fit an eyeball to the eye image, yielding improvements in gaze estimation.
We demonstrate that our method improves upon the state-of-the art in a number of tasks including eyelid registration and iris localization, in cross-dataset (person-independent) and personalized gaze estimation.
We show that both our model-based and feature-based methods out-perform an AlexNet baseline which has significantly more model parameters. We also show that the compact model can be used for robust real-time gaze estimation.
For the case of personalization we show that our landmarks-based SVR prevails especially in the case of few calibration samples.

\paragraph{Directions for future work.}
In our work, we provided initial insights on how \emph{explicitly} learned landmark-based features can out-perform \emph{implicitly} learned features in certain cases.
However, more research is necessary to understand which representation yields the most accurate and robust gaze estimation method.
We showed that our neural network is capable of detecting irises and eyelids to a higher accuracy than prior work.
However, direct evaluation of eye region landmark localization on current real-world datasets is challenging due to the absence of high-quality labels.
We experimentally showed encouraging performance gains for feature-based and model-based methods when leveraging our learned landmarks-based representation such as in the most challenging cross-dataset evaluation, contradicting previous comparisons \cite{Zhang2017PAMI}.
This implies that more research into feature-based and model-based methods could further improve gaze-estimation when coupled with powerful feature representations.

One particularly interesting result presented here is the capability to personalize deep-learning based gaze estimation methods, with our \emph{explicit} landmark-based feature performing very well for low number of calibration samples.
In future work this insight could be coupled with an appropriate online calibration procedure to produce highly accurate gaze estimates even in situations where only a user-facing camera is available and under difficult environmental conditions.
However, much future research is necessary to fully understand the best ways of garnering and utilizing calibration samples while improving gaze estimation accuracy significantly.
The hope is that our work will eventually enable conducting experiments at large scale by leveraging the commonly available cameras on devices such as smartphones, tablets, and laptops only.

\section*{Acknowledgements}
We would like to thank Erroll Wood, Tadas Baltru{\u s}aitis, and Wolfgang Fuhl for their help.
This work was supported in part by ERC Grant OPTINT (StG-2016-717054),
the Cluster of Excellence on Multimodal Computing and Interaction at Saarland University, Germany,
and by a JST CREST research grant (JPMJCR14E1), Japan.

\clearpage

\balance
\bibliographystyle{ACM-Reference-Format}
\bibliography{bibliography}


\begin{thebibliography}{53}


\ifx \showCODEN    \undefined \def \showCODEN     #1{\unskip}     \fi
\ifx \showDOI      \undefined \def \showDOI       #1{#1}\fi
\ifx \showISBNx    \undefined \def \showISBNx     #1{\unskip}     \fi
\ifx \showISBNxiii \undefined \def \showISBNxiii  #1{\unskip}     \fi
\ifx \showISSN     \undefined \def \showISSN      #1{\unskip}     \fi
\ifx \showLCCN     \undefined \def \showLCCN      #1{\unskip}     \fi
\ifx \shownote     \undefined \def \shownote      #1{#1}          \fi
\ifx \showarticletitle \undefined \def \showarticletitle #1{#1}   \fi
\ifx \showURL      \undefined \def \showURL       {\relax}        \fi
\providecommand\bibfield[2]{#2}
\providecommand\bibinfo[2]{#2}
\providecommand\natexlab[1]{#1}
\providecommand\showeprint[2][]{arXiv:#2}

\bibitem[\protect\citeauthoryear{Arar and Thiran}{Arar and Thiran}{2017}]%
        {Arar17arXiv}
\bibfield{author}{\bibinfo{person}{Nuri~Murat Arar} {and}
  \bibinfo{person}{Jean{-}Philippe Thiran}.} \bibinfo{year}{2017}\natexlab{}.
\newblock \showarticletitle{Robust Real-Time Multi-View Eye Tracking}.
\newblock \bibinfo{journal}{{\em CoRR Arxiv preprint\/}}
  \bibinfo{volume}{abs/1711.05444} (\bibinfo{year}{2017}).
\newblock
\showeprint[arxiv]{1711.05444}
\showURL{%
\url{http://arxiv.org/abs/1711.05444}}


\bibitem[\protect\citeauthoryear{Baluja and Pomerleau}{Baluja and
  Pomerleau}{1994}]%
        {Baluja1994}
\bibfield{author}{\bibinfo{person}{Shumeet Baluja} {and} \bibinfo{person}{Dean
  Pomerleau}.} \bibinfo{year}{1994}\natexlab{}.
\newblock \bibinfo{booktitle}{{\em Non-Intrusive Gaze Tracking Using Artificial
  Neural Networks}}.
\newblock \bibinfo{type}{{T}echnical {R}eport}. \bibinfo{address}{Pittsburgh,
  PA, USA}.
\newblock


\bibitem[\protect\citeauthoryear{Bengio, Louradour, Collobert, and
  Weston}{Bengio et~al\mbox{.}}{2009}]%
        {Bengio2009ICML}
\bibfield{author}{\bibinfo{person}{Yoshua Bengio},
  \bibinfo{person}{J{\'e}r\^{o}me Louradour}, \bibinfo{person}{Ronan
  Collobert}, {and} \bibinfo{person}{Jason Weston}.}
  \bibinfo{year}{2009}\natexlab{}.
\newblock \showarticletitle{Curriculum Learning}. In \bibinfo{booktitle}{{\em
  Proceedings of the 26th Annual International Conference on Machine Learning}}
  {\em (\bibinfo{series}{ICML '09})}. \bibinfo{publisher}{ACM},
  \bibinfo{address}{New York, NY, USA}, \bibinfo{pages}{41--48}.
\newblock
\showISBNx{978-1-60558-516-1}
\showDOI{%
\url{https://doi.org/10.1145/1553374.1553380}}


\bibitem[\protect\citeauthoryear{Biedert, Buscher, Schwarz, Hees, and
  Dengel}{Biedert et~al\mbox{.}}{2010}]%
        {biedert2010text}
\bibfield{author}{\bibinfo{person}{Ralf Biedert}, \bibinfo{person}{Georg
  Buscher}, \bibinfo{person}{Sven Schwarz}, \bibinfo{person}{J{\"o}rn Hees},
  {and} \bibinfo{person}{Andreas Dengel}.} \bibinfo{year}{2010}\natexlab{}.
\newblock \showarticletitle{Text 2.0}. In \bibinfo{booktitle}{{\em CHI'10
  Extended Abstracts on Human Factors in Computing Systems}}. ACM,
  \bibinfo{pages}{4003--4008}.
\newblock


\bibitem[\protect\citeauthoryear{Fuhl, Geisler, Santini, Rosenstiel, and
  Kasneci}{Fuhl et~al\mbox{.}}{2016a}]%
        {Fuhl2016Ubicomp}
\bibfield{author}{\bibinfo{person}{Wolfgang Fuhl}, \bibinfo{person}{David
  Geisler}, \bibinfo{person}{Thiago Santini}, \bibinfo{person}{Wolfgang
  Rosenstiel}, {and} \bibinfo{person}{Enkelejda Kasneci}.}
  \bibinfo{year}{2016}\natexlab{a}.
\newblock \showarticletitle{Evaluation of State-of-the-art Pupil Detection
  Algorithms on Remote Eye Images}. In \bibinfo{booktitle}{{\em Proceedings of
  the 2016 ACM International Joint Conference on Pervasive and Ubiquitous
  Computing: Adjunct}} {\em (\bibinfo{series}{UbiComp '16})}.
  \bibinfo{publisher}{ACM}, \bibinfo{address}{New York, NY, USA},
  \bibinfo{pages}{1716--1725}.
\newblock
\showISBNx{978-1-4503-4462-3}
\showDOI{%
\url{https://doi.org/10.1145/2968219.2968340}}


\bibitem[\protect\citeauthoryear{Fuhl, K{\"u}bler, Sippel, Rosenstiel, and
  Kasneci}{Fuhl et~al\mbox{.}}{2015}]%
        {Fuhl2015CAIP}
\bibfield{author}{\bibinfo{person}{Wolfgang Fuhl}, \bibinfo{person}{Thomas
  K{\"u}bler}, \bibinfo{person}{Katrin Sippel}, \bibinfo{person}{Wolfgang
  Rosenstiel}, {and} \bibinfo{person}{Enkelejda Kasneci}.}
  \bibinfo{year}{2015}\natexlab{}.
\newblock \bibinfo{booktitle}{{\em ExCuSe: Robust Pupil Detection in Real-World
  Scenarios}}.
\newblock \bibinfo{publisher}{Springer International Publishing},
  \bibinfo{address}{Cham}, \bibinfo{pages}{39--51}.
\newblock
\showISBNx{978-3-319-23192-1}
\showDOI{%
\url{https://doi.org/10.1007/978-3-319-23192-1_4}}


\bibitem[\protect\citeauthoryear{Fuhl, Santini, K\"{u}bler, and Kasneci}{Fuhl
  et~al\mbox{.}}{2016b}]%
        {Fuhl2016ETRA}
\bibfield{author}{\bibinfo{person}{Wolfgang Fuhl}, \bibinfo{person}{Thiago~C.
  Santini}, \bibinfo{person}{Thomas K\"{u}bler}, {and}
  \bibinfo{person}{Enkelejda Kasneci}.} \bibinfo{year}{2016}\natexlab{b}.
\newblock \showarticletitle{ElSe: Ellipse Selection for Robust Pupil Detection
  in Real-world Environments}. In \bibinfo{booktitle}{{\em Proceedings of the
  Ninth Biennial ACM Symposium on Eye Tracking Research \& Applications}} {\em
  (\bibinfo{series}{ETRA '16})}. \bibinfo{publisher}{ACM},
  \bibinfo{address}{New York, NY, USA}, \bibinfo{pages}{123--130}.
\newblock
\showISBNx{978-1-4503-4125-7}
\showDOI{%
\url{https://doi.org/10.1145/2857491.2857505}}


\bibitem[\protect\citeauthoryear{Funes~Mora, Monay, and Odobez}{Funes~Mora
  et~al\mbox{.}}{2014}]%
        {FunesMora2014ETRA}
\bibfield{author}{\bibinfo{person}{Kenneth~Alberto Funes~Mora},
  \bibinfo{person}{Florent Monay}, {and} \bibinfo{person}{Jean-Marc Odobez}.}
  \bibinfo{year}{2014}\natexlab{}.
\newblock \showarticletitle{EYEDIAP: A Database for the Development and
  Evaluation of Gaze Estimation Algorithms from RGB and RGB-D Cameras}. In
  \bibinfo{booktitle}{{\em Proceedings of the Symposium on Eye Tracking
  Research and Applications}} {\em (\bibinfo{series}{ETRA '14})}.
  \bibinfo{publisher}{ACM}, \bibinfo{address}{New York, NY, USA},
  \bibinfo{pages}{255--258}.
\newblock
\showISBNx{978-1-4503-2751-0}
\showDOI{%
\url{https://doi.org/10.1145/2578153.2578190}}


\bibitem[\protect\citeauthoryear{Funes-Mora and Odobez}{Funes-Mora and
  Odobez}{2016}]%
        {FunesMora2016IJCV}
\bibfield{author}{\bibinfo{person}{Kenneth~A. Funes-Mora} {and}
  \bibinfo{person}{Jean-Marc Odobez}.} \bibinfo{year}{2016}\natexlab{}.
\newblock \showarticletitle{Gaze Estimation in the 3D Space Using RGB-D
  Sensors}.
\newblock \bibinfo{journal}{{\em International Journal of Computer Vision\/}}
  \bibinfo{volume}{118}, \bibinfo{number}{2} (\bibinfo{date}{01 Jun}
  \bibinfo{year}{2016}), \bibinfo{pages}{194--216}.
\newblock
\showISSN{1573-1405}
\showDOI{%
\url{https://doi.org/10.1007/s11263-015-0863-4}}


\bibitem[\protect\citeauthoryear{Honari, Molchanov, Tyree, Vincent, Pal, and
  Kautz}{Honari et~al\mbox{.}}{2018}]%
        {Honari2018CVPR}
\bibfield{author}{\bibinfo{person}{Sina Honari}, \bibinfo{person}{Pavlo
  Molchanov}, \bibinfo{person}{Stephen Tyree}, \bibinfo{person}{Pascal
  Vincent}, \bibinfo{person}{Christopher Pal}, {and} \bibinfo{person}{Jan
  Kautz}.} \bibinfo{year}{2018}\natexlab{}.
\newblock \showarticletitle{Improving Landmark Localization with
  Semi-Supervised Learning}. In \bibinfo{booktitle}{{\em The IEEE Conference on
  Computer Vision and Pattern Recognition (CVPR)}}.
\newblock


\bibitem[\protect\citeauthoryear{Huang, Cai, Liu, Ahuja, and Zhang}{Huang
  et~al\mbox{.}}{2014a}]%
        {Huang14ETRA}
\bibfield{author}{\bibinfo{person}{Jia{-}Bin Huang}, \bibinfo{person}{Qin Cai},
  \bibinfo{person}{Zicheng Liu}, \bibinfo{person}{Narendra Ahuja}, {and}
  \bibinfo{person}{Zhengyou Zhang}.} \bibinfo{year}{2014}\natexlab{a}.
\newblock \showarticletitle{Towards accurate and robust cross-ratio based gaze
  trackers through learning from simulation}. In \bibinfo{booktitle}{{\em Eye
  Tracking Research and Applications, {ETRA} '14, Safety Harbor, FL, USA, March
  26-28, 2014}}. \bibinfo{pages}{75--82}.
\newblock
\showDOI{%
\url{https://doi.org/10.1145/2578153.2578162}}


\bibitem[\protect\citeauthoryear{Huang, Kwok, Ngai, Leong, and Chan}{Huang
  et~al\mbox{.}}{2014b}]%
        {Huang2014MM}
\bibfield{author}{\bibinfo{person}{Michael~Xuelin Huang},
  \bibinfo{person}{Tiffany~C.K. Kwok}, \bibinfo{person}{Grace Ngai},
  \bibinfo{person}{Hong~Va Leong}, {and} \bibinfo{person}{Stephen~C.F. Chan}.}
  \bibinfo{year}{2014}\natexlab{b}.
\newblock \showarticletitle{Building a Self-Learning Eye Gaze Model from User
  Interaction Data}. In \bibinfo{booktitle}{{\em Proceedings of the 22Nd ACM
  International Conference on Multimedia}} {\em (\bibinfo{series}{MM '14})}.
  \bibinfo{publisher}{ACM}, \bibinfo{address}{New York, NY, USA},
  \bibinfo{pages}{1017--1020}.
\newblock
\showISBNx{978-1-4503-3063-3}
\showDOI{%
\url{https://doi.org/10.1145/2647868.2655031}}


\bibitem[\protect\citeauthoryear{Huang, Veeraraghavan, and Sabharwal}{Huang
  et~al\mbox{.}}{2017}]%
        {Huang2017MVA}
\bibfield{author}{\bibinfo{person}{Qiong Huang}, \bibinfo{person}{Ashok
  Veeraraghavan}, {and} \bibinfo{person}{Ashutosh Sabharwal}.}
  \bibinfo{year}{2017}\natexlab{}.
\newblock \showarticletitle{TabletGaze: Dataset and Analysis for Unconstrained
  Appearance-based Gaze Estimation in Mobile Tablets}.
\newblock \bibinfo{journal}{{\em Mach. Vision Appl.\/}} \bibinfo{volume}{28},
  \bibinfo{number}{5-6} (\bibinfo{date}{Aug.} \bibinfo{year}{2017}),
  \bibinfo{pages}{445--461}.
\newblock
\showISSN{0932-8092}
\showDOI{%
\url{https://doi.org/10.1007/s00138-017-0852-4}}


\bibitem[\protect\citeauthoryear{Ioffe and Szegedy}{Ioffe and Szegedy}{2015}]%
        {Ioffe15ICML}
\bibfield{author}{\bibinfo{person}{Sergey Ioffe} {and}
  \bibinfo{person}{Christian Szegedy}.} \bibinfo{year}{2015}\natexlab{}.
\newblock \showarticletitle{Batch Normalization: Accelerating Deep Network
  Training by Reducing Internal Covariate Shift}. In \bibinfo{booktitle}{{\em
  Proceedings of the 32Nd International Conference on International Conference
  on Machine Learning - Volume 37}} {\em (\bibinfo{series}{ICML'15})}.
  \bibinfo{publisher}{JMLR.org}, \bibinfo{pages}{448--456}.
\newblock
\showURL{%
\url{http://dl.acm.org/citation.cfm?id=3045118.3045167}}


\bibitem[\protect\citeauthoryear{King}{King}{2009}]%
        {dlib}
\bibfield{author}{\bibinfo{person}{Davis~E. King}.}
  \bibinfo{year}{2009}\natexlab{}.
\newblock \showarticletitle{Dlib-ml: A Machine Learning Toolkit}.
\newblock \bibinfo{journal}{{\em Journal of Machine Learning Research\/}}
  \bibinfo{volume}{10} (\bibinfo{year}{2009}), \bibinfo{pages}{1755--1758}.
\newblock


\bibitem[\protect\citeauthoryear{Kingma and Ba}{Kingma and Ba}{2014}]%
        {Kingma2014arXiv}
\bibfield{author}{\bibinfo{person}{Diederik~P. Kingma} {and}
  \bibinfo{person}{Jimmy Ba}.} \bibinfo{year}{2014}\natexlab{}.
\newblock \showarticletitle{Adam: {A} Method for Stochastic Optimization}.
\newblock \bibinfo{journal}{{\em CoRR\/}}  \bibinfo{volume}{abs/1412.6980}
  (\bibinfo{year}{2014}).
\newblock
\showeprint[arxiv]{1412.6980}
\showURL{%
\url{http://arxiv.org/abs/1412.6980}}


\bibitem[\protect\citeauthoryear{Krafka, Khosla, Kellnhofer, Kannan,
  Bhandarkar, Matusik, and Torralba}{Krafka et~al\mbox{.}}{2016}]%
        {Krafka2016CVPR}
\bibfield{author}{\bibinfo{person}{Kyle Krafka}, \bibinfo{person}{Aditya
  Khosla}, \bibinfo{person}{Petr Kellnhofer}, \bibinfo{person}{Harini Kannan},
  \bibinfo{person}{Suchendra Bhandarkar}, \bibinfo{person}{Wojciech Matusik},
  {and} \bibinfo{person}{Antonio Torralba}.} \bibinfo{year}{2016}\natexlab{}.
\newblock \showarticletitle{Eye Tracking for Everyone}. In
  \bibinfo{booktitle}{{\em The IEEE Conference on Computer Vision and Pattern
  Recognition (CVPR)}}.
\newblock


\bibitem[\protect\citeauthoryear{Kunze, Bulling, Utsumi, Yuki, and Kise}{Kunze
  et~al\mbox{.}}{2013}]%
        {kunze13_iswc}
\bibfield{author}{\bibinfo{person}{Kai Kunze}, \bibinfo{person}{Andreas
  Bulling}, \bibinfo{person}{Yuzuko Utsumi}, \bibinfo{person}{Shiga Yuki},
  {and} \bibinfo{person}{Koichi Kise}.} \bibinfo{year}{2013}\natexlab{}.
\newblock \showarticletitle{I know what you are reading -- Recognition of
  document types using mobile eye tracking}. In \bibinfo{booktitle}{{\em Proc.
  IEEE International Symposium on Wearable Computers (ISWC)}}.
  \bibinfo{pages}{113--116}.
\newblock
\showDOI{%
\url{https://doi.org/10.1145/2493988.2494354}}


\bibitem[\protect\citeauthoryear{Li, Winfield, and Parkhurst}{Li
  et~al\mbox{.}}{2005}]%
        {Li2005CVPR}
\bibfield{author}{\bibinfo{person}{Dongheng Li}, \bibinfo{person}{David
  Winfield}, {and} \bibinfo{person}{Derrick~J. Parkhurst}.}
  \bibinfo{year}{2005}\natexlab{}.
\newblock \showarticletitle{Starburst: A Hybrid Algorithm for Video-based Eye
  Tracking Combining Feature-based and Model-based Approaches}. In
  \bibinfo{booktitle}{{\em Proceedings of the 2005 IEEE Computer Society
  Conference on Computer Vision and Pattern Recognition (CVPR'05) - Workshops -
  Volume 03}} {\em (\bibinfo{series}{CVPR '05})}. \bibinfo{publisher}{IEEE
  Computer Society}, \bibinfo{address}{Washington, DC, USA},
  \bibinfo{pages}{79--}.
\newblock
\showISBNx{0-7695-2372-2-3}
\showDOI{%
\url{https://doi.org/10.1109/CVPR.2005.531}}


\bibitem[\protect\citeauthoryear{Lu, Okabe, Sugano, and Sato}{Lu
  et~al\mbox{.}}{2011a}]%
        {Lu2011BMVC}
\bibfield{author}{\bibinfo{person}{Feng Lu}, \bibinfo{person}{Takahiro Okabe},
  \bibinfo{person}{Yusuke Sugano}, {and} \bibinfo{person}{Yoichi Sato}.}
  \bibinfo{year}{2011}\natexlab{a}.
\newblock \showarticletitle{A Head Pose-free Approach for Appearance-based Gaze
  Estimation}. In \bibinfo{booktitle}{{\em Proceedings of the British Machine
  Vision Conference}}. \bibinfo{publisher}{BMVA Press},
  \bibinfo{pages}{126.1--126.11}.
\newblock
\showISBNx{1-901725-43-X}
\newblock
\shownote{http://dx.doi.org/10.5244/C.25.126.}


\bibitem[\protect\citeauthoryear{Lu, Sugano, Okabe, and Sato}{Lu
  et~al\mbox{.}}{2011b}]%
        {Lu2011ICCV}
\bibfield{author}{\bibinfo{person}{Feng Lu}, \bibinfo{person}{Yusuke Sugano},
  \bibinfo{person}{Takahiro Okabe}, {and} \bibinfo{person}{Yoichi Sato}.}
  \bibinfo{year}{2011}\natexlab{b}.
\newblock \showarticletitle{Inferring Human Gaze from Appearance via Adaptive
  Linear Regression}. In \bibinfo{booktitle}{{\em Proceedings of the 2011
  International Conference on Computer Vision}} {\em (\bibinfo{series}{ICCV
  '11})}. \bibinfo{publisher}{IEEE Computer Society},
  \bibinfo{address}{Washington, DC, USA}, \bibinfo{pages}{153--160}.
\newblock
\showISBNx{978-1-4577-1101-5}
\showDOI{%
\url{https://doi.org/10.1109/ICCV.2011.6126237}}


\bibitem[\protect\citeauthoryear{Majaranta and Bulling}{Majaranta and
  Bulling}{2014}]%
        {majaranta14_apc}
\bibfield{author}{\bibinfo{person}{P\"aivi Majaranta} {and}
  \bibinfo{person}{Andreas Bulling}.} \bibinfo{year}{2014}\natexlab{}.
\newblock \bibinfo{booktitle}{{\em Eye Tracking and Eye-Based Human-Computer
  Interaction}}.
\newblock \bibinfo{publisher}{Springer}, \bibinfo{pages}{39--65}.
\newblock
\showISBNx{978-1-4471-6391-6}


\bibitem[\protect\citeauthoryear{Newell, Yang, and Deng}{Newell
  et~al\mbox{.}}{2016}]%
        {Newell2016ECCV}
\bibfield{author}{\bibinfo{person}{Alejandro Newell}, \bibinfo{person}{Kaiyu
  Yang}, {and} \bibinfo{person}{Jia Deng}.} \bibinfo{year}{2016}\natexlab{}.
\newblock \showarticletitle{Stacked hourglass networks for human pose
  estimation}. In \bibinfo{booktitle}{{\em European Conference on Computer
  Vision}}. Springer, \bibinfo{pages}{483--499}.
\newblock


\bibitem[\protect\citeauthoryear{Sagonas, Tzimiropoulos, Zafeiriou, and
  Pantic}{Sagonas et~al\mbox{.}}{2013}]%
        {Sagonas2013ICCVW}
\bibfield{author}{\bibinfo{person}{Christos Sagonas}, \bibinfo{person}{Georgios
  Tzimiropoulos}, \bibinfo{person}{Stefanos Zafeiriou}, {and}
  \bibinfo{person}{Maja Pantic}.} \bibinfo{year}{2013}\natexlab{}.
\newblock \showarticletitle{300 faces in-the-wild challenge: The first facial
  landmark localization challenge}. In \bibinfo{booktitle}{{\em Proceedings of
  the IEEE International Conference on Computer Vision Workshops}}.
  \bibinfo{pages}{397--403}.
\newblock


\bibitem[\protect\citeauthoryear{Sesma, Villanueva, and Cabeza}{Sesma
  et~al\mbox{.}}{2012}]%
        {Sesma2012ETRA}
\bibfield{author}{\bibinfo{person}{Laura Sesma}, \bibinfo{person}{Arantxa
  Villanueva}, {and} \bibinfo{person}{Rafael Cabeza}.}
  \bibinfo{year}{2012}\natexlab{}.
\newblock \showarticletitle{Evaluation of Pupil Center-eye Corner Vector for
  Gaze Estimation Using a Web Cam}. In \bibinfo{booktitle}{{\em Proceedings of
  the Symposium on Eye Tracking Research and Applications}} {\em
  (\bibinfo{series}{ETRA '12})}. \bibinfo{publisher}{ACM},
  \bibinfo{address}{New York, NY, USA}, \bibinfo{pages}{217--220}.
\newblock
\showISBNx{978-1-4503-1221-9}
\showDOI{%
\url{https://doi.org/10.1145/2168556.2168598}}


\bibitem[\protect\citeauthoryear{Shrivastava, Pfister, Tuzel, Susskind, Wang,
  and Webb}{Shrivastava et~al\mbox{.}}{2017}]%
        {Shrivastava2017CVPR}
\bibfield{author}{\bibinfo{person}{Ashish Shrivastava}, \bibinfo{person}{Tomas
  Pfister}, \bibinfo{person}{Oncel Tuzel}, \bibinfo{person}{Joshua Susskind},
  \bibinfo{person}{Wenda Wang}, {and} \bibinfo{person}{Russell Webb}.}
  \bibinfo{year}{2017}\natexlab{}.
\newblock \showarticletitle{Learning From Simulated and Unsupervised Images
  Through Adversarial Training}. In \bibinfo{booktitle}{{\em The IEEE
  Conference on Computer Vision and Pattern Recognition (CVPR)}}.
\newblock


\bibitem[\protect\citeauthoryear{Smith, Yin, Feiner, and Nayar}{Smith
  et~al\mbox{.}}{2013}]%
        {Smith2013UIST}
\bibfield{author}{\bibinfo{person}{Brian~A. Smith}, \bibinfo{person}{Qi Yin},
  \bibinfo{person}{Steven~K. Feiner}, {and} \bibinfo{person}{Shree~K. Nayar}.}
  \bibinfo{year}{2013}\natexlab{}.
\newblock \showarticletitle{Gaze Locking: Passive Eye Contact Detection for
  Human-object Interaction}. In \bibinfo{booktitle}{{\em Proceedings of the
  26th Annual ACM Symposium on User Interface Software and Technology}} {\em
  (\bibinfo{series}{UIST '13})}. \bibinfo{publisher}{ACM},
  \bibinfo{address}{New York, NY, USA}, \bibinfo{pages}{271--280}.
\newblock
\showISBNx{978-1-4503-2268-3}
\showDOI{%
\url{https://doi.org/10.1145/2501988.2501994}}


\bibitem[\protect\citeauthoryear{Sugano, Matsushita, and Sato}{Sugano
  et~al\mbox{.}}{2014}]%
        {Sugano2014CVPR}
\bibfield{author}{\bibinfo{person}{Yusuke Sugano}, \bibinfo{person}{Yasuyuki
  Matsushita}, {and} \bibinfo{person}{Yoichi Sato}.}
  \bibinfo{year}{2014}\natexlab{}.
\newblock \showarticletitle{Learning-by-Synthesis for Appearance-Based 3D Gaze
  Estimation}. In \bibinfo{booktitle}{{\em 2014 IEEE Conference on Computer
  Vision and Pattern Recognition}}. \bibinfo{pages}{1821--1828}.
\newblock
\showISSN{1063-6919}
\showDOI{%
\url{https://doi.org/10.1109/CVPR.2014.235}}


\bibitem[\protect\citeauthoryear{Sugano, Zhang, and Bulling}{Sugano
  et~al\mbox{.}}{2016}]%
        {Sugano2016UIST}
\bibfield{author}{\bibinfo{person}{Yusuke Sugano}, \bibinfo{person}{Xucong
  Zhang}, {and} \bibinfo{person}{Andreas Bulling}.}
  \bibinfo{year}{2016}\natexlab{}.
\newblock \showarticletitle{AggreGaze: Collective Estimation of Audience
  Attention on Public Displays}. In \bibinfo{booktitle}{{\em Proceedings of the
  29th Annual Symposium on User Interface Software and Technology}} {\em
  (\bibinfo{series}{UIST '16})}. \bibinfo{publisher}{ACM},
  \bibinfo{address}{New York, NY, USA}, \bibinfo{pages}{821--831}.
\newblock
\showISBNx{978-1-4503-4189-9}
\showDOI{%
\url{https://doi.org/10.1145/2984511.2984536}}


\bibitem[\protect\citeauthoryear{Sun, Liu, and Sun}{Sun et~al\mbox{.}}{2015}]%
        {Sun2015Elsevier}
\bibfield{author}{\bibinfo{person}{Li Sun}, \bibinfo{person}{Zicheng Liu},
  {and} \bibinfo{person}{Ming-Ting Sun}.} \bibinfo{year}{2015}\natexlab{}.
\newblock \showarticletitle{Real Time Gaze Estimation with a Consumer Depth
  Camera}.
\newblock \bibinfo{journal}{{\em Inf. Sci.\/}} \bibinfo{volume}{320},
  \bibinfo{number}{C} (\bibinfo{date}{Nov.} \bibinfo{year}{2015}),
  \bibinfo{pages}{346--360}.
\newblock
\showISSN{0020-0255}
\showDOI{%
\url{https://doi.org/10.1016/j.ins.2015.02.004}}


\bibitem[\protect\citeauthoryear{Sun, Wang, and Tang}{Sun
  et~al\mbox{.}}{2013}]%
        {Sun2013CVPR}
\bibfield{author}{\bibinfo{person}{Yi Sun}, \bibinfo{person}{Xiaogang Wang},
  {and} \bibinfo{person}{Xiaoou Tang}.} \bibinfo{year}{2013}\natexlab{}.
\newblock \showarticletitle{Deep Convolutional Network Cascade for Facial Point
  Detection}. In \bibinfo{booktitle}{{\em The IEEE Conference on Computer
  Vision and Pattern Recognition (CVPR)}}.
\newblock


\bibitem[\protect\citeauthoryear{Tan, Kriegman, and Ahuja}{Tan
  et~al\mbox{.}}{2002}]%
        {Tan02WACV}
\bibfield{author}{\bibinfo{person}{Kar-Han Tan}, \bibinfo{person}{David~J.
  Kriegman}, {and} \bibinfo{person}{Narendra Ahuja}.}
  \bibinfo{year}{2002}\natexlab{}.
\newblock \showarticletitle{Appearance-based Eye Gaze Estimation}. In
  \bibinfo{booktitle}{{\em Proceedings of the Sixth IEEE Workshop on
  Applications of Computer Vision}} {\em (\bibinfo{series}{WACV '02})}.
  \bibinfo{publisher}{IEEE Computer Society}, \bibinfo{address}{Washington, DC,
  USA}, \bibinfo{pages}{191--}.
\newblock
\showISBNx{0-7695-1858-3}
\showURL{%
\url{http://dl.acm.org/citation.cfm?id=832302.836853}}


\bibitem[\protect\citeauthoryear{Timm and Barth}{Timm and Barth}{2011}]%
        {Timm2011VISAPP}
\bibfield{author}{\bibinfo{person}{Fabian Timm} {and} \bibinfo{person}{Erhardt
  Barth}.} \bibinfo{year}{2011}\natexlab{}.
\newblock \showarticletitle{Accurate Eye Centre Localisation by Means of
  Gradients.}. In \bibinfo{booktitle}{{\em VISAPP}}.
  \bibinfo{publisher}{SciTePress}, \bibinfo{pages}{125--130}.
\newblock
\showISBNx{978-989-8425-47-8}
\showURL{%
\url{http://dblp.uni-trier.de/db/conf/visapp/visapp2011.html#TimmB11}}


\bibitem[\protect\citeauthoryear{Tompson, Jain, LeCun, and Bregler}{Tompson
  et~al\mbox{.}}{2014}]%
        {Tompson2014NIPS}
\bibfield{author}{\bibinfo{person}{Jonathan Tompson}, \bibinfo{person}{Arjun
  Jain}, \bibinfo{person}{Yann LeCun}, {and} \bibinfo{person}{Christoph
  Bregler}.} \bibinfo{year}{2014}\natexlab{}.
\newblock \showarticletitle{Joint Training of a Convolutional Network and a
  Graphical Model for Human Pose Estimation}. In \bibinfo{booktitle}{{\em
  Proceedings of the 27th International Conference on Neural Information
  Processing Systems - Volume 1}} {\em (\bibinfo{series}{NIPS'14})}.
  \bibinfo{publisher}{MIT Press}, \bibinfo{address}{Cambridge, MA, USA},
  \bibinfo{pages}{1799--1807}.
\newblock
\showURL{%
\url{http://dl.acm.org/citation.cfm?id=2968826.2969027}}


\bibitem[\protect\citeauthoryear{Toshev and Szegedy}{Toshev and
  Szegedy}{2014}]%
        {Toshev2014CVPR}
\bibfield{author}{\bibinfo{person}{Alexander Toshev} {and}
  \bibinfo{person}{Christian Szegedy}.} \bibinfo{year}{2014}\natexlab{}.
\newblock \showarticletitle{DeepPose: Human Pose Estimation via Deep Neural
  Networks}. In \bibinfo{booktitle}{{\em Proceedings of the 2014 IEEE
  Conference on Computer Vision and Pattern Recognition}} {\em
  (\bibinfo{series}{CVPR '14})}. \bibinfo{publisher}{IEEE Computer Society},
  \bibinfo{address}{Washington, DC, USA}, \bibinfo{pages}{1653--1660}.
\newblock
\showISBNx{978-1-4799-5118-5}
\showDOI{%
\url{https://doi.org/10.1109/CVPR.2014.214}}


\bibitem[\protect\citeauthoryear{Valenti, Sebe, and Gevers}{Valenti
  et~al\mbox{.}}{2012}]%
        {Valenti2012TIP}
\bibfield{author}{\bibinfo{person}{Roberto Valenti}, \bibinfo{person}{Nicu
  Sebe}, {and} \bibinfo{person}{Theo Gevers}.} \bibinfo{year}{2012}\natexlab{}.
\newblock \showarticletitle{Combining Head Pose and Eye Location Information
  for Gaze Estimation}.
\newblock \bibinfo{journal}{{\em Trans. Img. Proc.\/}} \bibinfo{volume}{21},
  \bibinfo{number}{2} (\bibinfo{date}{Feb.} \bibinfo{year}{2012}),
  \bibinfo{pages}{802--815}.
\newblock
\showISSN{1057-7149}
\showDOI{%
\url{https://doi.org/10.1109/TIP.2011.2162740}}


\bibitem[\protect\citeauthoryear{Villanueva, Ponz, Sesma-Sanchez, Ariz, Porta,
  and Cabeza}{Villanueva et~al\mbox{.}}{2013}]%
        {Villanueva2013TOMM}
\bibfield{author}{\bibinfo{person}{Arantxa Villanueva},
  \bibinfo{person}{Victoria Ponz}, \bibinfo{person}{Laura Sesma-Sanchez},
  \bibinfo{person}{Mikel Ariz}, \bibinfo{person}{Sonia Porta}, {and}
  \bibinfo{person}{Rafael Cabeza}.} \bibinfo{year}{2013}\natexlab{}.
\newblock \showarticletitle{Hybrid Method Based on Topography for Robust
  Detection of Iris Center and Eye Corners}.
\newblock \bibinfo{journal}{{\em ACM Trans. Multimedia Comput. Commun.
  Appl.\/}} \bibinfo{volume}{9}, \bibinfo{number}{4}, Article
  \bibinfo{articleno}{25} (\bibinfo{date}{Aug.} \bibinfo{year}{2013}),
  \bibinfo{numpages}{20}~pages.
\newblock
\showISSN{1551-6857}
\showDOI{%
\url{https://doi.org/10.1145/2501643.2501647}}


\bibitem[\protect\citeauthoryear{Wang, Sung, and Venkateswarlu}{Wang
  et~al\mbox{.}}{2003}]%
        {Wang2003ICCV}
\bibfield{author}{\bibinfo{person}{Jian-Gang Wang}, \bibinfo{person}{Eric
  Sung}, {and} \bibinfo{person}{Ronda Venkateswarlu}.}
  \bibinfo{year}{2003}\natexlab{}.
\newblock \showarticletitle{Eye gaze estimation from a single image of one
  eye}. In \bibinfo{booktitle}{{\em Computer Vision, 2003. Proceedings. Ninth
  IEEE International Conference on}}. IEEE, \bibinfo{pages}{136--143}.
\newblock


\bibitem[\protect\citeauthoryear{Wang and Ji}{Wang and Ji}{2017}]%
        {Wang2017ICCV}
\bibfield{author}{\bibinfo{person}{Kang Wang} {and} \bibinfo{person}{Qiang
  Ji}.} \bibinfo{year}{2017}\natexlab{}.
\newblock \showarticletitle{Real Time Eye Gaze Tracking with 3D Deformable
  Eye-Face Model}. In \bibinfo{booktitle}{{\em Proceedings of the 2017 IEEE
  International Conference on Computer Vision (ICCV)}} {\em
  (\bibinfo{series}{ICCV '17})}. \bibinfo{publisher}{IEEE Computer Society},
  \bibinfo{address}{Washington, DC, USA}.
\newblock


\bibitem[\protect\citeauthoryear{Wood, Baltruaitis, Zhang, Sugano, Robinson,
  and Bulling}{Wood et~al\mbox{.}}{2015}]%
        {Wood2015ICCV}
\bibfield{author}{\bibinfo{person}{Erroll Wood}, \bibinfo{person}{Tadas
  Baltruaitis}, \bibinfo{person}{Xucong Zhang}, \bibinfo{person}{Yusuke
  Sugano}, \bibinfo{person}{Peter Robinson}, {and} \bibinfo{person}{Andreas
  Bulling}.} \bibinfo{year}{2015}\natexlab{}.
\newblock \showarticletitle{Rendering of Eyes for Eye-Shape Registration and
  Gaze Estimation}. In \bibinfo{booktitle}{{\em Proceedings of the 2015 IEEE
  International Conference on Computer Vision (ICCV)}} {\em
  (\bibinfo{series}{ICCV '15})}. \bibinfo{publisher}{IEEE Computer Society},
  \bibinfo{address}{Washington, DC, USA}, \bibinfo{pages}{3756--3764}.
\newblock
\showISBNx{978-1-4673-8391-2}
\showDOI{%
\url{https://doi.org/10.1109/ICCV.2015.428}}


\bibitem[\protect\citeauthoryear{Wood, Baltru{\v{s}}aitis, Morency, Robinson,
  and Bulling}{Wood et~al\mbox{.}}{2016a}]%
        {Wood2016ECCV}
\bibfield{author}{\bibinfo{person}{Erroll Wood}, \bibinfo{person}{Tadas
  Baltru{\v{s}}aitis}, \bibinfo{person}{Louis-Philippe Morency},
  \bibinfo{person}{Peter Robinson}, {and} \bibinfo{person}{Andreas Bulling}.}
  \bibinfo{year}{2016}\natexlab{a}.
\newblock \showarticletitle{A 3D morphable eye region model for gaze
  estimation}. In \bibinfo{booktitle}{{\em European Conference on Computer
  Vision}}. Springer, \bibinfo{pages}{297--313}.
\newblock


\bibitem[\protect\citeauthoryear{Wood, Baltru\v{s}aitis, Morency, Robinson, and
  Bulling}{Wood et~al\mbox{.}}{2016b}]%
        {Wood2016ETRA}
\bibfield{author}{\bibinfo{person}{Erroll Wood}, \bibinfo{person}{Tadas
  Baltru\v{s}aitis}, \bibinfo{person}{Louis-Philippe Morency},
  \bibinfo{person}{Peter Robinson}, {and} \bibinfo{person}{Andreas Bulling}.}
  \bibinfo{year}{2016}\natexlab{b}.
\newblock \showarticletitle{Learning an Appearance-based Gaze Estimator from
  One Million Synthesised Images}. In \bibinfo{booktitle}{{\em Proceedings of
  the Ninth Biennial ACM Symposium on Eye Tracking Research \& Applications}}
  {\em (\bibinfo{series}{ETRA '16})}. \bibinfo{publisher}{ACM},
  \bibinfo{address}{New York, NY, USA}, \bibinfo{pages}{131--138}.
\newblock
\showISBNx{978-1-4503-4125-7}
\showDOI{%
\url{https://doi.org/10.1145/2857491.2857492}}


\bibitem[\protect\citeauthoryear{Wood and Bulling}{Wood and Bulling}{2014}]%
        {Wood2014ETRAEA}
\bibfield{author}{\bibinfo{person}{Erroll Wood} {and} \bibinfo{person}{Andreas
  Bulling}.} \bibinfo{year}{2014}\natexlab{}.
\newblock \showarticletitle{EyeTab: Model-based Gaze Estimation on Unmodified
  Tablet Computers}. In \bibinfo{booktitle}{{\em Proceedings of the Symposium
  on Eye Tracking Research and Applications}} {\em (\bibinfo{series}{ETRA
  '14})}. \bibinfo{publisher}{ACM}, \bibinfo{address}{New York, NY, USA},
  \bibinfo{pages}{207--210}.
\newblock
\showISBNx{978-1-4503-2751-0}
\showDOI{%
\url{https://doi.org/10.1145/2578153.2578185}}


\bibitem[\protect\citeauthoryear{Xiong, Liu, Cai, and Zhang}{Xiong
  et~al\mbox{.}}{2014}]%
        {Xiong2014Ubicomp}
\bibfield{author}{\bibinfo{person}{Xuehan Xiong}, \bibinfo{person}{Zicheng
  Liu}, \bibinfo{person}{Qin Cai}, {and} \bibinfo{person}{Zhengyou Zhang}.}
  \bibinfo{year}{2014}\natexlab{}.
\newblock \showarticletitle{Eye Gaze Tracking Using an RGBD Camera: A
  Comparison with a RGB Solution}. In \bibinfo{booktitle}{{\em Proceedings of
  the 2014 ACM International Joint Conference on Pervasive and Ubiquitous
  Computing: Adjunct Publication}} {\em (\bibinfo{series}{UbiComp '14
  Adjunct})}. \bibinfo{publisher}{ACM}, \bibinfo{address}{New York, NY, USA},
  \bibinfo{pages}{1113--1121}.
\newblock
\showISBNx{978-1-4503-3047-3}
\showDOI{%
\url{https://doi.org/10.1145/2638728.2641694}}


\bibitem[\protect\citeauthoryear{Xu, Ehinger, Zhang, Finkelstein, Kulkarni, and
  Xiao}{Xu et~al\mbox{.}}{2015}]%
        {xu2015turkergaze}
\bibfield{author}{\bibinfo{person}{Pingmei Xu}, \bibinfo{person}{Krista~A
  Ehinger}, \bibinfo{person}{Yinda Zhang}, \bibinfo{person}{Adam Finkelstein},
  \bibinfo{person}{Sanjeev~R Kulkarni}, {and} \bibinfo{person}{Jianxiong
  Xiao}.} \bibinfo{year}{2015}\natexlab{}.
\newblock \showarticletitle{Turkergaze: Crowdsourcing saliency with webcam
  based eye tracking}.
\newblock \bibinfo{journal}{{\em arXiv preprint arXiv:1504.06755\/}}
  (\bibinfo{year}{2015}).
\newblock


\bibitem[\protect\citeauthoryear{Yang, Liu, and Zhang}{Yang
  et~al\mbox{.}}{2017}]%
        {Yang2017CVPRW}
\bibfield{author}{\bibinfo{person}{Jing Yang}, \bibinfo{person}{Qingshan Liu},
  {and} \bibinfo{person}{Kaihua Zhang}.} \bibinfo{year}{2017}\natexlab{}.
\newblock \showarticletitle{Stacked hourglass network for robust facial
  landmark localisation}. In \bibinfo{booktitle}{{\em Computer Vision and
  Pattern Recognition Workshops (CVPRW), 2017 IEEE Conference on}}. IEEE,
  \bibinfo{pages}{2025--2033}.
\newblock


\bibitem[\protect\citeauthoryear{Yoo, Lee, and Chung}{Yoo
  et~al\mbox{.}}{2002}]%
        {Yoo02}
\bibfield{author}{\bibinfo{person}{Dong~Hyun Yoo}, \bibinfo{person}{Bang~Rae
  Lee}, {and} \bibinfo{person}{Myoung~Jin Chung}.}
  \bibinfo{year}{2002}\natexlab{}.
\newblock \showarticletitle{Non-Contact Eye Gaze Tracking System by Mapping of
  Corneal Reflections}. In \bibinfo{booktitle}{{\em Proceedings of the Fifth
  IEEE International Conference on Automatic Face and Gesture Recognition}}
  {\em (\bibinfo{series}{FGR '02})}. \bibinfo{publisher}{IEEE Computer
  Society}, \bibinfo{address}{Washington, DC, USA}, \bibinfo{pages}{101--}.
\newblock
\showISBNx{0-7695-1602-5}


\bibitem[\protect\citeauthoryear{Zafeiriou, Trigeorgis, Chrysos, Deng, and
  Shen}{Zafeiriou et~al\mbox{.}}{2017}]%
        {Zafeiriou2017CVPRW}
\bibfield{author}{\bibinfo{person}{Stefanos Zafeiriou}, \bibinfo{person}{George
  Trigeorgis}, \bibinfo{person}{Grigorios Chrysos}, \bibinfo{person}{Jiankang
  Deng}, {and} \bibinfo{person}{Jie Shen}.} \bibinfo{year}{2017}\natexlab{}.
\newblock \showarticletitle{The Menpo Facial Landmark Localisation Challenge: A
  Step Towards the Solution}. In \bibinfo{booktitle}{{\em The IEEE Conference
  on Computer Vision and Pattern Recognition (CVPR) Workshops}}.
\newblock


\bibitem[\protect\citeauthoryear{Zhang, Sugano, Fritz, and Bulling}{Zhang
  et~al\mbox{.}}{2015}]%
        {Zhang2015CVPR}
\bibfield{author}{\bibinfo{person}{Xucong Zhang}, \bibinfo{person}{Yusuke
  Sugano}, \bibinfo{person}{Mario Fritz}, {and} \bibinfo{person}{Andreas
  Bulling}.} \bibinfo{year}{2015}\natexlab{}.
\newblock \showarticletitle{Appearance-Based Gaze Estimation in the Wild}. In
  \bibinfo{booktitle}{{\em The IEEE Conference on Computer Vision and Pattern
  Recognition (CVPR)}}. \bibinfo{pages}{4511--4520}.
\newblock
\showDOI{%
\url{https://doi.org/10.1109/CVPR.2015.7299081}}


\bibitem[\protect\citeauthoryear{Zhang, Sugano, Fritz, and Bulling}{Zhang
  et~al\mbox{.}}{2017}]%
        {Zhang2017CVPRW}
\bibfield{author}{\bibinfo{person}{Xucong Zhang}, \bibinfo{person}{Yusuke
  Sugano}, \bibinfo{person}{Mario Fritz}, {and} \bibinfo{person}{Andreas
  Bulling}.} \bibinfo{year}{2017}\natexlab{}.
\newblock \showarticletitle{It's Written All Over Your Face: Full-Face
  Appearance-Based Gaze Estimation}. In \bibinfo{booktitle}{{\em The IEEE
  Conference on Computer Vision and Pattern Recognition (CVPR) Workshops}}.
  \bibinfo{pages}{2299--2308}.
\newblock
\showDOI{%
\url{https://doi.org/10.1109/CVPRW.2017.284}}


\bibitem[\protect\citeauthoryear{Zhang, Sugano, Fritz, and Bulling}{Zhang
  et~al\mbox{.}}{2018}]%
        {Zhang2017PAMI}
\bibfield{author}{\bibinfo{person}{Xucong Zhang}, \bibinfo{person}{Yusuke
  Sugano}, \bibinfo{person}{Mario Fritz}, {and} \bibinfo{person}{Andreas
  Bulling}.} \bibinfo{year}{2018}\natexlab{}.
\newblock \showarticletitle{MPIIGaze: Real-World Dataset and Deep
  Appearance-Based Gaze Estimation}.
\newblock \bibinfo{journal}{{\em IEEE Transactions on Pattern Analysis and
  Machine Intelligence\/}} (\bibinfo{year}{2018}).
\newblock
\showDOI{%
\url{https://doi.org/10.1109/TPAMI.2017.2778103}}


\bibitem[\protect\citeauthoryear{Zhang, Bulling, and Gellersen}{Zhang
  et~al\mbox{.}}{2013}]%
        {zhang13_chi}
\bibfield{author}{\bibinfo{person}{Yanxia Zhang}, \bibinfo{person}{Andreas
  Bulling}, {and} \bibinfo{person}{Hans Gellersen}.}
  \bibinfo{year}{2013}\natexlab{}.
\newblock \showarticletitle{SideWays: A Gaze Interface for Spontaneous
  Interaction with Situated Displays}. In \bibinfo{booktitle}{{\em Proc. of the
  31st SIGCHI International Conference on Human Factors in Computing Systems
  (CHI 2013)}} (2013-04-27). \bibinfo{publisher}{ACM}, \bibinfo{address}{New
  York, NY, USA}, \bibinfo{pages}{851--860}.
\newblock
\showISBNx{978-1-4503-1899-0}


\bibitem[\protect\citeauthoryear{Zhang, Bulling, and Gellersen}{Zhang
  et~al\mbox{.}}{2014}]%
        {Zhang14_AVIb}
\bibfield{author}{\bibinfo{person}{Yanxia Zhang}, \bibinfo{person}{Andreas
  Bulling}, {and} \bibinfo{person}{Hans Gellersen}.}
  \bibinfo{year}{2014}\natexlab{}.
\newblock \showarticletitle{Pupil-canthi-ratio: a calibration-free method for
  tracking horizontal gaze direction}. In \bibinfo{booktitle}{{\em Proc. of the
  2014 International Working Conference on Advanced Visual Interfaces (AVI
  14)}} (2014-05-27). \bibinfo{publisher}{ACM}, \bibinfo{address}{New York, NY,
  USA}, \bibinfo{pages}{129--132}.
\newblock


\end{thebibliography}

\end{document}